\RequirePackage{iftex}
\ifXeTeX
\PassOptionsToPackage{xetex}{hyperref}
\AddToHook{package/microtype/after}{\microtypesetup{tracking=false}\renewcommand{\DisableLigatures}[2][]{}}
\fi
\pdfoutput=1
\documentclass[]{TEAI}
\ifXeTeX
\fi
\usepackage{array}
\usepackage{tabularx}
\usepackage{ragged2e}
\usepackage{float}
\usepackage{wrapfig}
\usepackage{needspace}
\usepackage{enumitem}
\usepackage{makecell}
\usepackage{nicefrac}
\usepackage{xcolor}
\usepackage[export]{adjustbox}

\setcitestyle{authoryear}

\usepackage[utf8]{inputenc}
\ifPDFTeX
\DeclareUnicodeCharacter{2191}{\ensuremath{\uparrow}}
\DeclareUnicodeCharacter{2193}{\ensuremath{\downarrow}}
\else
\usepackage{newunicodechar}
\newunicodechar{↑}{\ensuremath{\uparrow}}
\newunicodechar{↓}{\ensuremath{\downarrow}}
\fi
\definecolor{user}{RGB}{30,100,180}
\definecolor{assistant}{RGB}{180,80,20}
\definecolor{tool}{RGB}{20,130,60}
\definecolor{system}{RGB}{120,50,150}   
\title{Environments as Scaffold: Enriching Feedback to Bootstrap Self-Evolving Agents in Long-Horizon Tasks}

\author{
    Hongbang Yuan\textsuperscript{1},
    Zhuoran Jin\textsuperscript{2},
    Yixin Cao\textsuperscript{1,3,$\dagger$}
}

\affiliation[1]{\mbox{Fudan University}}
\affiliation[2]{\mbox{CASIA}} 
\affiliation[3]{\mbox{Shanghai Innovation Institute}}

\abstract{ 
    Large Language Models demonstrate remarkable proficiency in static reasoning, yet training them as autonomous agents through Reinforcement Learning (RL) for long-horizon tasks is often hindered by severe reward sparsity. While conventional \textit{agent-side warming} up via supervised fine-tuning (SFT) can alleviate this, it is frequently limited by data scarcity and constrained exploration. To address this, we propose a paradigm shift to  \textit{environment-side adaptation} by constructing \textbf{F}eedback-\textbf{E}nriched \textbf{E}nvironments (\textbf{FEEs}). Through a pilot study, we establish a feedback design strategy that reformulates environments by transitioning from action guidance to observation enrichment during the later stages of both intra-episode exploration and inter-episode evolution. Large-scale experiments on SciWorld and BFCL benchmarks using various Qwen3 model scales and RL algorithms such as GRPO, GSPO, and DAPO demonstrate that FEEs consistently yield performance improvements over standard settings. Furthermore, our analysis reveals that training with FEEs \textbf{(1)} stabilizes training dynamics by reducing entropy volatility,  \textbf{(2)} facilitates proactive state-space exploration in difficult tasks, \textbf{(3) }ensures the internalization of environmental guidance into policy weights rather than acting as a mere inference-time prior, and \textbf{(4) }identifies intra-group feedback consistency as a critical boundary for stable optimization. 
    }

\correspondence{\email{hbyuan25@m.fudan.edu.cn, yxcao@fudan.edu.cn}}
\checkdata[Code]{\url{https://github.com/HongbangYuan/EnvAsScaffold}}

\begin{document}
\maketitle

\begingroup
\renewcommand{\thefootnote}{} 
\footnotetext{$\dagger$ Corresponding author.}
\endgroup

\vspace{-1.0em}

\section{Introduction}

Large Language Models (LLMs) \citep{qwen3,gpt4,llama3,deepseekr1} have demonstrated remarkable proficiency in domains such as  mathematical reasoning \citep{glazer2024frontiermath,du2025nemotron-math} and code generation \citep{yang2024swe-agent,jimenez2024swe-bench}. As the field advances, the research focus is shifting from solving these \textit{static, single-turn tasks} to building autonomous agents capable of exploring \textit{dynamic, complex real-world environments}, such as web navigation \citep{zhou2024webarena,bai2026webgym} and terminal-based computer usage \citep{merrill2026terminal-bench,gandhi2026endlessTerminals}. These tasks are inherently \textbf{long-horizon} \citep{zhang2025rlvmr,00332025agentLearningEarly}, often necessitating sequences of dozens of interaction steps to achieve a solution. 
To solve them, reinforcement learning (RL) based approaches  \citep{chen2025scalingAgentLearning,feng2025group-in-group,wang2025harnessingUncertainty} have emerged as a central direction, where agents learn to adapt their policies through iterative feedback from the environments.

However, RL on long-horizon tasks suffers greatly from \textbf{reward sparsity}. Specifically, the agent lacks the capability for effective exploration and is prone to getting trapped in zero-reward trajectories, leading to vanishing gradients that leave the agent with no signal to learn from.  While warming up the agent via supervised fine-tuning (SFT) on expert trajectories effectively increases the probability of obtaining initial positive rewards \citep{wen2025light-r1CurriculumSft,00012025acereason-nemotron11}, it faces significant bottlenecks. Not only is the collection of ground-truth trajectories expensive and hard to scale \citep{chen2025reinforcementLearningLong-horizon}, but over-optimizing the SFT objective also risks overly constraining the agent's behavior, thereby restricting the exploration potential necessary for effective RL \citep{kang2025quagmiresSft-rlPost-training}.

\begin{figure*}[t]
    \centering
    \includegraphics[width=\linewidth]{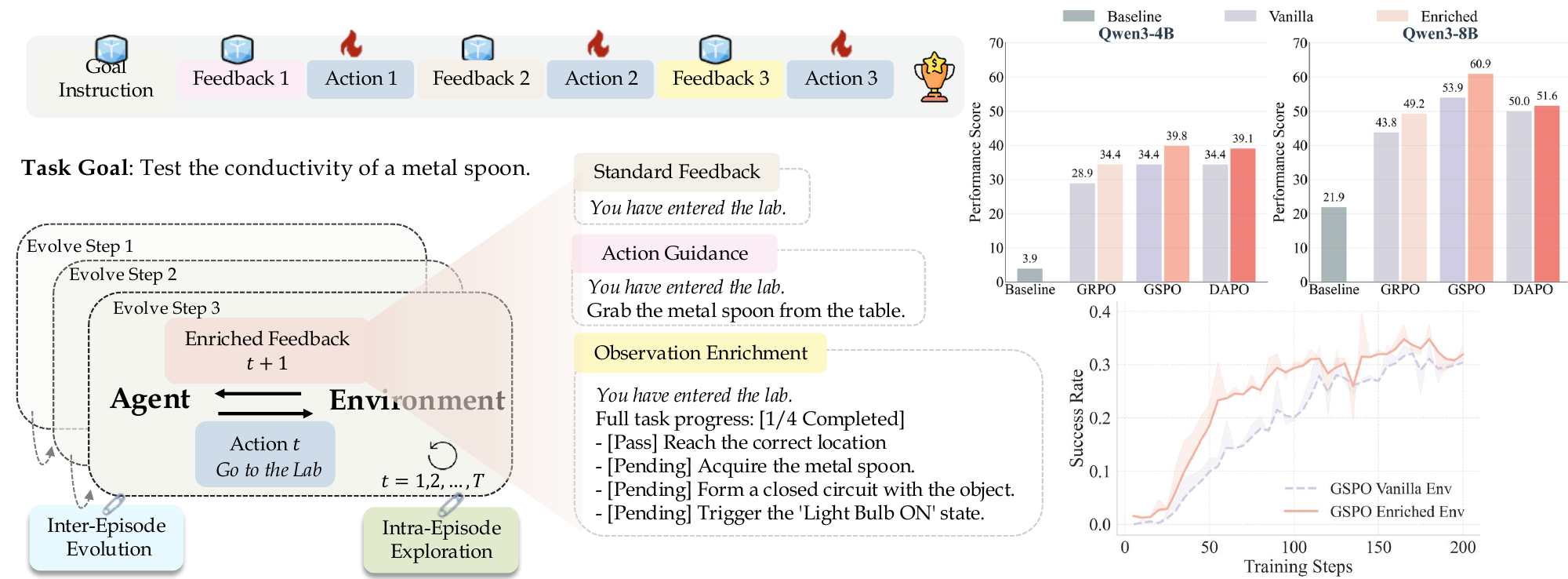}
    \caption{Illustration of the Feedback-Enriched Environments (FEEs).}
    \label{fig:introduction}
\end{figure*}

To address this, we propose a paradigm shift from \textbf{agent-side} warming up to \textbf{environment-side} adaptation. Specifically, instead of seeking a better-initialized agent, we propose to construct more suitable environments during RL by systematically diversifying and enriching their feedback signals.
As shown in Figure \ref{fig:introduction}, while standard feedback leaves the agent searching blindly, enriched feedback guides it to enter the lab and retrieve the metal spoon, which boosts the success rate and helps the agent better understand the environment.

In particular, we study two key dimensions of environment feedback design: \textbf{what} information to provide and \textbf{when} to deliver it to the agent. For the former, we consider \textit{action guidance} for suggesting immediate next steps and \textit{observation enrichment} for providing supplementary state information. For the latter, we analyze when these signals are presented across two temporal scales: \textit{intra-episode exploration} within a single trajectory and \textit{inter-episode evolution} throughout the training lifecycle. Our initial study suggests that the most effective enriching strategy is to deliver action guidance during early interaction steps and training phases, and then transition to observation enrichment in the later stages of both processes.

To systematically validate this strategy, we build \textbf{F}eedback-\textbf{E}nriched \textbf{E}nvironments (\textbf{FEEs}) based on standard environments from two widely adopted benchmarks: SciWorld \citep{wang2022scienceworld} and BFCL \citep{patil2025berkeleyFunctionCallingLeaderboard}. Within these FEEs, we train Qwen3-4B and Qwen3-8B models with various agent-side RL  algorithms, including GRPO \citep{shao2024deepseekmathPushingLimits}, DAPO \citep{yu2025dapoOpen-sourceLlm}, and GSPO \citep{zheng2025groupSequencePolicy}. 
Extensive empirical results demonstrate that training in FEEs consistently outperforms training in standard environments, yielding an average improvement of $2.82\%$ across various model scales and RL algorithms.

Additionally, to  gain insights beyond performance scores, we further analyze the impacts of training agents with our proposed feedback-enriched environments. 
\textbf{(1)} \textit{How do FEEs contribute to agent-side RL training stability?} Through an analysis of policy entropy dynamics, we find that FEEs act as a stabilizer that reduces gradient volatility and ensures smoother convergence, a benefit that persists even under explicit entropy regularization.  
\textbf{(2)} \textit{Do FEEs promote more effective state-space exploration?} By analyzing average success rates during training and accuracy across environments of varying difficulty levels, we find that that FEEs enable agents to access previously unreachable states while retaining robust exploration capabilities in standard environments. 
\textbf{(3)} \textit{Is the environmental feedback internalized into the agent's policy weights?} Our analysis shows that feedback is internalized into the policy weights rather than acting as a mere inference-time prior. 
\textbf{(4)} \textit{Should environmental feedback remain consistent or diverse within sampling groups?} We investigate the boundary of feedback diversification and find that intra-group consistency is crucial for stable optimization, as excessive stochasticity within a single rollout group introduces harmful noise.

 Our contributions can be summarized as follows:

\begin{itemize}
    \item We propose a paradigm shift from agent-side training to environment-side adaptation. We develop a systematic feedback design strategy that optimizes what information to provide (action guidance vs. observation enrichment) and when to provide it (intra-episode vs. inter-episode) to facilitate more effective RL training in long-horizon tasks.

    \item Based on our strategy, we construct FEEs using the SciWorld and BFCL benchmarks. Extensive experiments across various model scales (Qwen3-4B/8B) and RL algorithms (GRPO, DAPO, GSPO) demonstrate that FEEs consistently improve performance comparing with the standard environments.

    \item  We provide an in-depth analysis of the impacts of training with FEEs. Our findings reveal that these environments promote proactive state-space exploration in difficult tasks, stabilize training dynamics through entropy reduction, ensure the internalization of external guidance into policy weights, and identify intra-group feedback consistency as a critical factor for stable optimization.

\end{itemize}

\section{Preliminary}
\label{sec:preliminary}

\paragraph{Environment Definition}
The environment can be defined as a goal-conditioned Partially Observable Markov Decision Process (POMDP):
\begin{equation}
    e_{\phi} = (\mathcal{G},  \mathcal{S},  \mathcal{A}, O, \mathcal{T}, r)
\end{equation}

where $\mathcal{G}$ is a set of potential goals in to be accomplished by the agent, $S$ is a set of internal states, $A$ is the set of actions available for the agent, $O$ is a set observations accessible to the agent, $\mathcal{T}:\mathcal{S} \times \mathcal{A} \rightarrow \mathcal{S} $ is the state transition probability function, and  $r: \mathcal{S} \times \mathcal{A} \times \mathcal{G}  \rightarrow  \mathbb{R} $ is the reward function.

\paragraph{Agent Exploration}
Formally, the agent $\pi_\theta$ interacts with the environment $e_\phi$ to achieve a goal $g$ within a finite horizon of $T$ discrete time steps. Given the current partial observation $o_t \in O$ and the interaction history $h_t$, the agent generates a textual action $a_t \in \mathcal{A}$. Accordingly, the agent's behavior is modeled as a conditional distribution over the output tokens, formally defined as:

{\small \begin{equation}
\label{agent_exploration_equation}
\begin{gathered}
a_t \sim \pi_\theta\Big(a_t \mid g, o_t,h_t\Big), \\
h_t = (o_1, a_1),(o_2,a_2), \dots, (o_{t-1}, a_{t-1})
\end{gathered}
\end{equation}}

Upon executing $a_t$, the environment transitions to the next state $s_{t+1}$ governed by the transition function $\mathcal{T}(s_{t+1} \mid s_t, a_t)$ and generates the subsequent observation $o_{t+1}$. The episode trajectory is represented as $\tau = (o_1, a_1, \dots, o_T, a_T)$, where the final performance is evaluated by a trajectory-level reward $r_\tau$, typically serving as a binary indicator of task success or failure. We define \textbf{enriched feedback} as the result of an intervention on the observation space, $o^+_t = \mathcal{E}(o_t, h_t)$, forming a new environment $e_{\phi}^+ = (\mathcal{G}, \mathcal{S}, \mathcal{A}, \Omega^+, \mathcal{T}, \mathcal{R})$. Consequently, the agent's behavior $a_t \sim \pi_\theta(a_t \mid g, o^+_t, h_t)$ is shifted, implicitly reshaping the action distribution to vary task difficulty and encourage trajectory diversity.
For clarity, we henceforth refer to the original environment  $e_\phi$  as the \textbf{standard environment} and $e_{\phi}^+$ as the \textbf{enriched environment}.

\paragraph{Agentic RL}

Given a goal $g$, the agent samples a batch of $N$ candidate trajectories $\{\tau_1, \tau_2, \dots, \tau_N\}$ according to its current policy $\pi_\theta$. Upon completion, each trajectory $\tau_k$ receives the final trajectory-level reward $r(\tau_k)$. The advantage $A(\tau_k)$ for each trajectory is computed based on the group statistics:
\begin{equation}
    A(\tau_k) = GroupComputation[(r_{\tau_k})_{k=1}^{N}]
\end{equation}
Subsequently, the computed trajectory-level advantage $A(\tau_k)$ is assigned uniformly to all \textbf{agent-generated tokens} $[a_i]_{i=1}^{T}$ within $\tau_k$. Notably, tokens corresponding to \textbf{environment-generated tokens} $[o_i]_{i=1}^{T}$ are masked out, ensuring that the policy gradient is estimated solely based on the agent's actions. We term the continuous refinement of the agent's policy $\pi_\theta$ via these gradients as \textbf{agent evolving}.
More details can be found in Appendix \ref{appendix:rl_algorithms}.

\section{Feedback Design Strategy for FEEs}
In this section, we empirically investigate how to design effective FEEs through the lens of \textbf{what} information to provide and \textbf{when} to provide it. Specifically, we categorize feedback into action guidance and observation enrichment, examining their roles during intra-episode exploration and inter-episode evolution. The results demonstrate that action guidance is better suited for the initial stages, while state enrichment is more beneficial for the later stages of both exploration and training.

\subsection{Design Choices}

\textbf{Action Guidance (AG).}  Action guidance integrates procedural hints into environmental feedback to effectively narrow the search space of the policy $\pi_\theta$. This acts as a soft intervention, preventing the agent from becoming trapped in redundant exploration loops. For instance, as shown in Figure \ref{fig:introduction}, in a conductivity experiment, an unguided agent might erroneously waste steps exploring a classroom for materials. Action guidance, however, explicitly prompts the agent to ``\textit{enter the laboratory}'', thereby pruning this irrelevant branch and steering the trajectory efficiently toward the goal.

\textbf{Observation Enrichment (OE).} Complementarily, observation enrichment augments raw observations with supplementary semantic information to address the inherent partial observability of the POMDP environment. For instance, in a circuit assembly task, while a generic observation merely state ``\textit{wire connected}'', enriched feedback elucidates hidden states, such as ``\textit{the cathode remains unpowered}''. This transparency empowers the agent to identify the missing link and execute remedial actions, rather than guessing blindly.

\textbf{Intra-episode Exploration.} This dimension corresponds to the step-wise interaction process within a single episode’s finite horizon $T$. Specifically, it investigates whether enriched feedback should be introduced during the initial stages of exploration or delayed until the later phases of the episode.

\textbf{Inter-episode Evolution.} This dimension tracks the continuous refinement of the agent's policy $\pi_\theta$ across the entire training lifecycle. Specifically, it investigates whether enriched feedback should be introduced in the early stages of evolution when the agent's capabilities are limited, or in the later stages when it has become more proficient.

\subsection{Empirical Validation}
\label{empirical_validation}

\textbf{Experimental Setup.}
To evaluate different feedback strategies, we construct enriched variants of the standard SciWorld environment \citep{wang2022scienceworld}, which evaluates the capability of agents to design and execute elementary science experiments within interactive text-based environments. Specifically, we limit each episode to 15 steps, where \textbf{AG-Early} and \textbf{AG-Late} provide action guidance at steps 1--3 and 6--10, respectively, while \textbf{OE-Early} and \textbf{OE-Late} introduce observation enrichment during the same intervals. Throughout the entire agent evolution process, each enrichment operation is applied stochastically with a 0.5 probability. More details can be found in Appendix \ref{appendix:enrichment_strategies}.

\textbf{Implementation}.
Training is conducted with \texttt{Qwen3-4B-Thinking-2507} using GRPO \citep{shao2024deepseekmathPushingLimits} for 200 steps across the four enriched environments. Each training step utilizes 16 parallel environments with a 15-interaction rollout length and a $1\times 10^{-6}$ learning rate. The task success rate is evaluated on the standard environment every 5 training steps, using a strictly disjoint set of tasks to prevent training contamination.

\Needspace{23\baselineskip}
\subsection{Results and Analysis}

\begin{wrapfigure}{r}{0.48\textwidth}
    \centering
    \includegraphics[width=\linewidth]{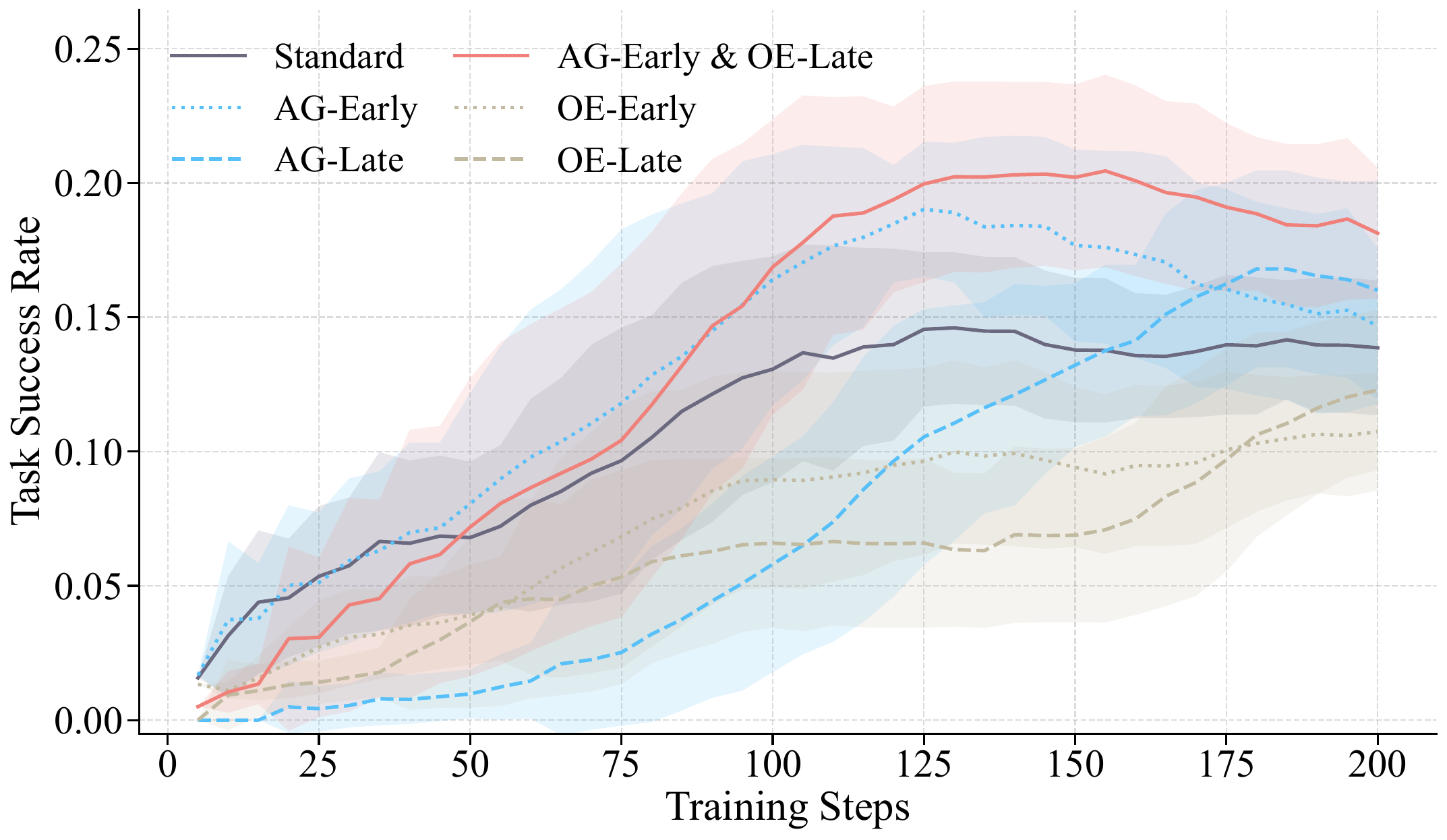}
    \caption{RL training dynamics under various feedback enrichment strategies. Task success rate is evaluated on the standard environment. Curves are smoothed for visual clarity, and shaded regions denote variance. }
    \label{fig:what_info_exp}
\end{wrapfigure}

\textbf{Action guidance generally outperforms observation enrichment.}
As shown in Figure \ref{fig:what_info_exp}, the blue curves  consistently maintain a superior performance margin over the brown curves throughout the training process. A plausible rationale is that action guidance explicitly prunes the massive search space by prescribing valid next steps, effectively bypassing the exploration bottleneck. In contrast, observation enrichment provides supplementary semantic information which imposes a higher cognitive load. The agent must implicitly learn to map these new features to optimal actions via complex causal reasoning, a process inherently slower than following procedural instructions.

\textbf{What we enrich defines when we enrich.} 
As shown in Figure \ref{fig:what_info_exp}, AG-Early outperforms AG-Late in earlier stages, while OE-Late exhibits a sharp upward trend in the later stages compared to OE-Early. This indicates that from both intra- and inter-episode perspectives, AG should be applied during the initial stage, while OE shoud be introduced in the later stage. A plausible explanation is that action guidance effectively prunes the initial combinatorial search space to bootstrap early learning, but becomes redundant once basic navigation is mastered. Conversely, observation enrichment imposes a higher cognitive load and requires a foundational policy to interpret the additional semantics, making it less effective initially but highly potent later on.

\textbf{Combining \textit{AG-Early} and \textit{OE-Late} yields superior performance.}
To further validate that distinct strategies suit different training stages, we introduce a hybrid approach applying AG-Early for the first 100 steps and switching to OE-Late for the remaining 100 steps. As shown by the red curve in Figure \ref{fig:what_info_exp}, this combination surpasses all other settings, achieving the highest success rate. This success highlights a general strategy for environment enrichment: \textit{across both the agent-environment interaction and agent evolution processes, action guidance should be leveraged during the early stage to bootstrap exploration, while observation enrichment should be introduced in the later stage for advanced policy refinement}.

\section{Experiments}

\begin{table*}[t!]
\centering
\resizebox{\textwidth}{!}{\begin{tabular}{rcccccc}
\toprule
\multicolumn{1}{c}{\multirow{2}{*}{\textbf{Model}}} & \multicolumn{4}{c}{\textbf{BFCL V3 Multi-Turn }} & \multirow{2}{*}{\textbf{SciWorld}} & \multirow{2}{*}{\textbf{Average}} \\ \cmidrule(lr){2-5}
 & \textbf{Base} & \textbf{Long Context} & \textbf{Miss Func} & \textbf{Miss Param} & & \\ \midrule

\rowcolor[HTML]{F5F5F5} \multicolumn{1}{l}{GPT-5.4} & 47.00 & 56.00 & 56.00 & 49.00 & 52.34 & 52.07 \\ 
\rowcolor[HTML]{F5F5F5}\multicolumn{1}{l}{Kimi-K2-Thinking} & 69.00 & 54.00 & 68.00 & 57.00 & 19.53 & 53.51 \\
\rowcolor[HTML]{F5F5F5} \multicolumn{1}{l}{Qwen3-235B-Thinking}  & 36.00 & 29.00 & 32.00 & 23.00 & 55.47 & 35.09 \\ 
\midrule
 
\multicolumn{1}{l}{\textit{\textbf{Qwen3-4B}}} & 35.00 & 28.00 & 27.00 & 24.00 & 3.90 & 23.58 \\ 

\rowcolor[HTML]{FFFBF1} GRPO + Standard & 58.00 & 52.00 & 34.00 & 32.00 & 28.91 & 40.98 \\
\rowcolor[HTML]{FFFBF1} GRPO + Enriched & 68.00 \textcolor[HTML]{006400}{($\uparrow$10.00\%)} & 46.00 \textcolor[HTML]{A52424}{($\downarrow$6.00\%)} & 43.00 \textcolor[HTML]{0A6803}{($\uparrow$9.00\%)} & 30.00 \textcolor[HTML]{BF4949}{($\downarrow$2.00\%)} & 34.38 \textcolor[HTML]{30770F}{($\uparrow$5.47\%)} & 44.27 \textcolor[HTML]{478017}{($\uparrow$3.29\%)} \\ 

\rowcolor[HTML]{FCF8F8} DAPO + Standard & 61.00 & 44.00 & 39.00 & 41.00 & 34.38 & 43.88 \\
\rowcolor[HTML]{FCF8F8} DAPO + Enriched & 71.00 \textcolor[HTML]{006400}{($\uparrow$10.00\%)} & 50.00 \textcolor[HTML]{2A740E}{($\uparrow$6.00\%)} & 45.00 \textcolor[HTML]{2A740E}{($\uparrow$6.00\%)} & 34.00 \textcolor[HTML]{9E1B1B}{($\downarrow$7.00\%)} & 39.06 \textcolor[HTML]{387A12}{($\uparrow$4.68\%)} & 47.81 \textcolor[HTML]{407D15}{($\uparrow$3.93\%)} \\ 

\rowcolor[HTML]{F2F9FF} GSPO + Standard & 54.00 & 40.00 & 29.00 & 25.00 & 34.38 & 36.48 \\
\rowcolor[HTML]{F2F9FF} GSPO + Enriched & 54.00 & 40.00 & 37.00 \textcolor[HTML]{156C07}{($\uparrow$8.00\%)} & 22.00 \textcolor[HTML]{B94040}{($\downarrow$3.00\%)} & 39.84 \textcolor[HTML]{30770F}{($\uparrow$5.46\%)} & 38.57 \textcolor[HTML]{54851B}{($\uparrow$2.09\%)} \\ 
\midrule
\multicolumn{1}{l}{\textit{\textbf{Qwen3-8B}}} & 35.00 & 24.00 & 30.00 & 25.00 & 21.88 & 27.18 \\ 
\rowcolor[HTML]{FFFBF1} GRPO + Standard & 66.00 & 35.00 & 52.00 & 37.00 & 43.75 & 46.75 \\
\rowcolor[HTML]{FFFBF1} GRPO + Enriched & 66.00 & 36.00 \textcolor[HTML]{60891F}{($\uparrow$1.00\%)} & 48.00 \textcolor[HTML]{B23737}{($\downarrow$4.00\%)} & 38.00 \textcolor[HTML]{60891F}{($\uparrow$1.00\%)} & 49.22 \textcolor[HTML]{30770F}{($\uparrow$5.47\%)} & 47.44 \textcolor[HTML]{638B20}{($\uparrow$0.69\%)} \\ 

\rowcolor[HTML]{FCF8F8} DAPO + Standard & 65.00 & 33.00 & 47.00 & 34.00 & 50.00 & 45.80 \\
\rowcolor[HTML]{FCF8F8} DAPO + Enriched & 71.00 \textcolor[HTML]{2A740E}{($\uparrow$6.00\%)} & 35.00 \textcolor[HTML]{55851C}{($\uparrow$2.00\%)} & 49.00 \textcolor[HTML]{55851C}{($\uparrow$2.00\%)} & 34.00 & 51.56 \textcolor[HTML]{5A871D}{($\uparrow$1.56\%)} & 48.11 \textcolor[HTML]{52841A}{($\uparrow$2.31\%)} \\ 

\rowcolor[HTML]{F2F9FF} GSPO + Standard & 58.00 & 29.00 & 41.00 & 27.00 & 53.91 & 41.78 \\
\rowcolor[HTML]{F2F9FF} GSPO + Enriched & 65.00 \textcolor[HTML]{20700A}{($\uparrow$7.00\%)} & 32.00 \textcolor[HTML]{4A8118}{($\uparrow$3.00\%)} & 40.00 \textcolor[HTML]{C65252}{($\downarrow$1.00\%)} & 34.00 \textcolor[HTML]{20700A}{($\uparrow$7.00\%)} & 60.94 \textcolor[HTML]{1F700A}{($\uparrow$7.03\%)} & 46.39 \textcolor[HTML]{397A12}{($\uparrow$4.61\%)} \\ 
\bottomrule
\end{tabular}}
\caption{Main evaluation results on two selected benchmarks. Color blocks group identical RL algorithms to show the impact of training environments. Performance scores are reported as percentages (\%). Values in parentheses indicate the absolute percentage point improvement (\textcolor[HTML]{006400}{↑}) or degradation (\textcolor[HTML]{C65252}{↓}) of FEE training over the standard baseline.}
\label{tab:main_exp_results}
\end{table*}

\subsection{Experiment Setting}

\textbf{Enriched Environments.}
To systematically validate our \textit{AG-Early and OE-Late} strategy, we conduct enriched environments for agent training from two primary benchmarks: SciWorld and BFCL-v3 Multi-Turn \citep{patil2025berkeleyFunctionCallingLeaderboard}, which is introduced to evaluate the agents' capability to handle dynamic and realistic user interactions across multiple dialogue turns via predefined APIs.  In SciWorld, action guidance represents concrete valid actions, and observation enrichment represents real-time task progress tracking. In BFCL, action guidance represents suggested actions appended to the user query, and observtion enrichments represents expanded raw return contents of invoked tools.
Consistent with Section \ref{empirical_validation}, we train the agent for 200 steps with a 0.5 enrichment probability, transitioning from AG-Early to OE-Late at step 100, while ensuring that the standard evaluation tasks remain strictly disjoint.
Detailed implementations and examples are provided in Appendix \ref{appendix:enrichment_strategies}.

\textbf{RL Algorithms.} We evaluate our approach with three representative RL algorithms: GRPO \citep{shao2024deepseekmathPushingLimits}, DAPO \citep{yu2025dapoOpen-sourceLlm}, and GSPO \citep{zheng2025groupSequencePolicy}. GRPO estimates the advantage directly by computing relative scores within a group, thereby bypassing the need for a separate critic model. DAPO incorporates specialized strategies such as clip-higher and dynamic sampling to stabilize optimization. GSPO calculates the importance ratio based on sequence-level likelihoods. Details of the mathematical expressions of the algorithms can be found in Appendix \ref{appendix:rl_algorithms}.

\textbf{Training.} We adopt the 4B and 8B variants of the Qwen3 series as the foundational backbones for RL training. Training utilizes a $1\times10^{-6}$ learning rate with evaluations conducted every 5 steps, where the peak performance is recorded in Table \ref{tab:main_exp_results}. To establish a performance ceiling, we incorporate leading proprietary models as baselines, including GPT-5.4, Kimi-K2-Thinking \citep{kimi_k2}, Qwen3-235B-Thinking \citep{qwen3}.

\subsection{Main Results}

\textbf{RL training effectively bridges the gap between open-source models and leading proprietary models.} RL training significantly enhances the model's capabilities in multi-turn interactions. As shown in Table \ref{tab:main_exp_results}, Qwen3-4B and Qwen3-8B models initially exhibit limited proficiency, yielding average scores of only $23.58\%$ and $27.18\%$, respectively. After RL training, Qwen3-4B and 8B reach $47.81\%$ and $48.11\%$, comfortably outperforming Qwen3-235B-Thinking and closely approaching top-tier closed-source models like GPT-5.4. This confirms that RL training effectively transforms foundational knowledge into specialized execution skills for complex agentic tasks.

\textbf{Training agents in FEEs consistently yields superior performance compared to standard environments.} This advantage remains robust across all evaluated model scales, optimization algorithms, and benchmarks. For instance, as shown in Table \ref{tab:main_exp_results}, the Qwen3-8B model optimized via GSPO on SciWorld improves from $53.91\%$ to $60.94\%$ when trained with enriched feedback. Similarly, the Qwen3-4B model using GRPO on BFCL-Base tasks achieves a $10.00\%$ absolute performance increase. These consistent gains validate the effectiveness of our feedback enrichment strategy.

\textbf{FEEs may lead agents to rely too much on environmental feedback, weakening their ability to question the input context.} Despite widespread gains, training with FEEs may cause slight performance regressions in scenarios where agents must question input sufficiency rather than blindly execute commands. As shown in Table \ref{tab:main_exp_results}, with FEEs, the Qwen3-4B model trained with DAPO drops from $41.00\%$ to $34.00\%$ on Miss Param tasks where essential information is missing from user requests. Similarly, the Qwen3-8B model with GRPO declines from $52.00\%$ to $48.00\%$ on Miss Func tasks where no available tools are provided to fulfill the user request. We hypothesize that the proactive guidance in FEEs makes agents overly inclined to follow interaction cues, leading them to prioritize execution over necessary clarification.

\section{Discussion}

\subsection{Training Stability}

\textbf{Exp 1.} 
To investigate how FEEs affect training stability, we track the evolution of policy entropy across four experimental configurations. Specifically, we compare standard environments and FEEs under two distinct settings: with and without an entropy regularization loss. Since entropy loss explicitly forces the agent to maintain high policy diversity for exploration, we examine whether FEEs provide independent stability even under such volatile conditions. The resulting entropy curves of Qwen3-4B over 300 training steps on SciWorld related environments are presented in Figure \ref{fig:stability_exp}.

\textbf{Res 1.}
As shown in Figure \ref{fig:stability_exp}, training within FEEs significantly delays or prevents premature policy collapse compared to standard environments. Specifically, in configurations without explicit entropy regularization represented by the solid lines, the agent trained in FEEs maintains steady entropy throughout the entire 300 steps without collapsing, whereas the standard environment suffers from a sharp drop to zero at approximately 250 steps. When entropy regularization is introduced to force exploration represented by the dashed lines, FEEs successfully sustain stable training for nearly 200 steps, while the standard environment collapses much earlier at around 130 steps.

\textbf{Takeaway 1.}  
FEEs stabilize RL training with more diverse environment feedback, regardless of explicit entropy regularization.

\begin{figure}[t]
    \begin{minipage}[t]{0.48\textwidth}
    \vspace{0pt}
    \centering
    \includegraphics[width=\linewidth]{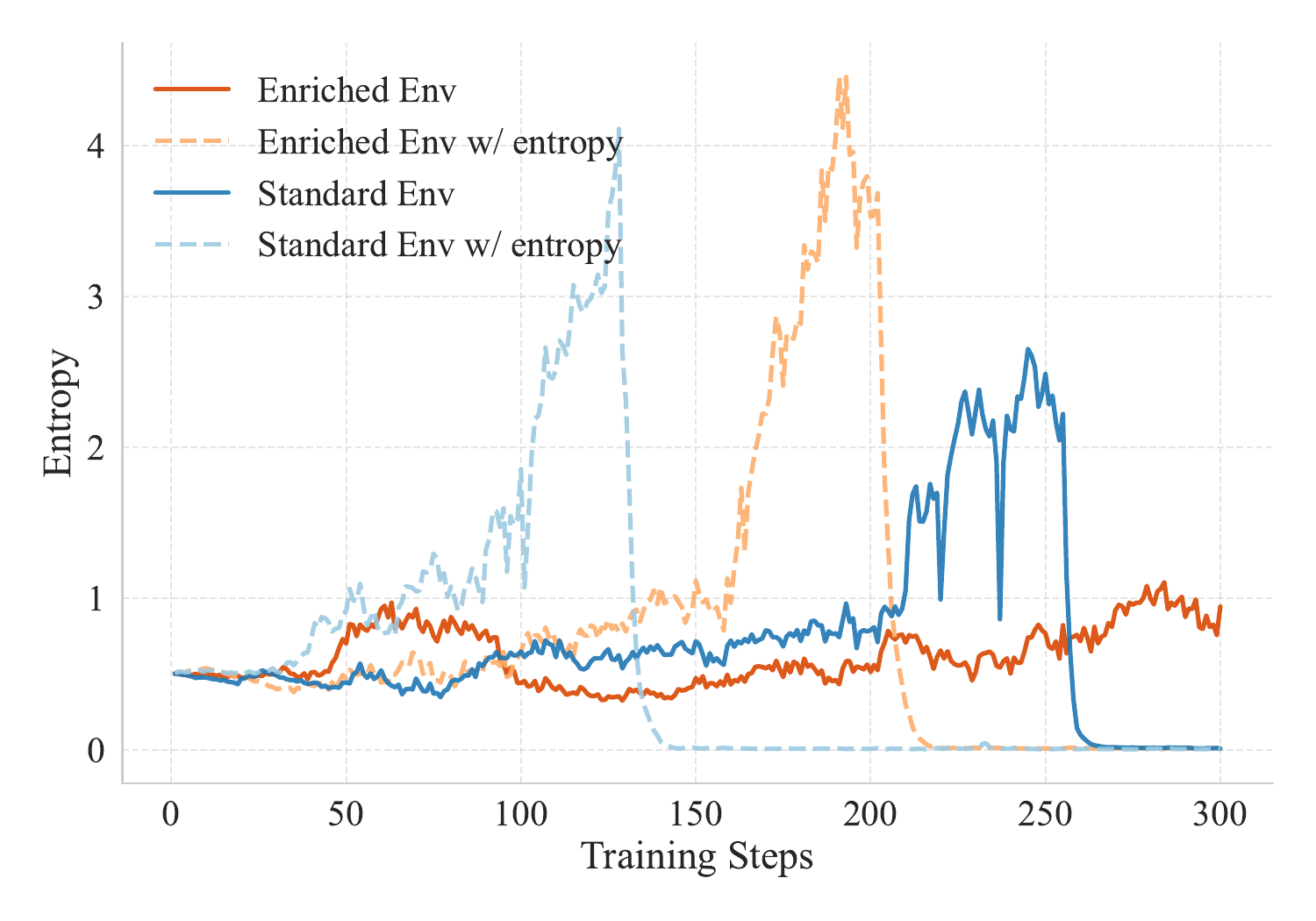}
    \caption{Policy entropy curves of Qwen3-4B over 300 training steps in standard SciWorld environments and FEEs, with and without entropy regularization.}
    \label{fig:stability_exp}
    \end{minipage}
    \hfill
    \begin{minipage}[t]{0.48\textwidth}
    \vspace{0pt}
    \centering
    \includegraphics[width=\linewidth]{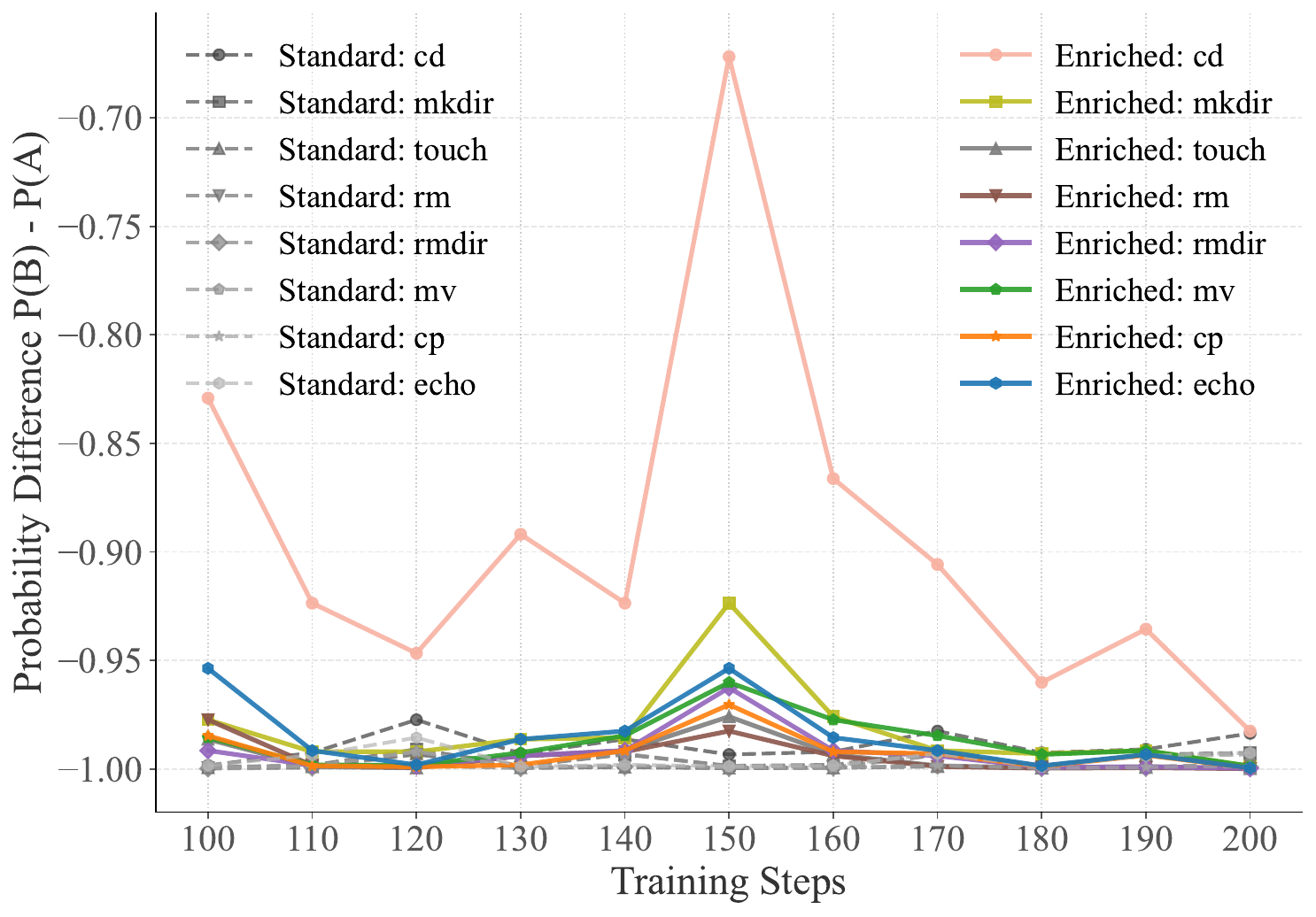}
    \setcounter{figure}{4}
    \caption{Probability difference between the original and enriched feedback options across training steps.}
    \label{fig:internalization_exp}
    \end{minipage}
    \setcounter{figure}{3}
\end{figure}

\subsection{State-Space Exploration}

\begin{figure*}[ht!]     \centering
        \begin{minipage}{0.66\linewidth}
        \centering
        \includegraphics[width=\linewidth]{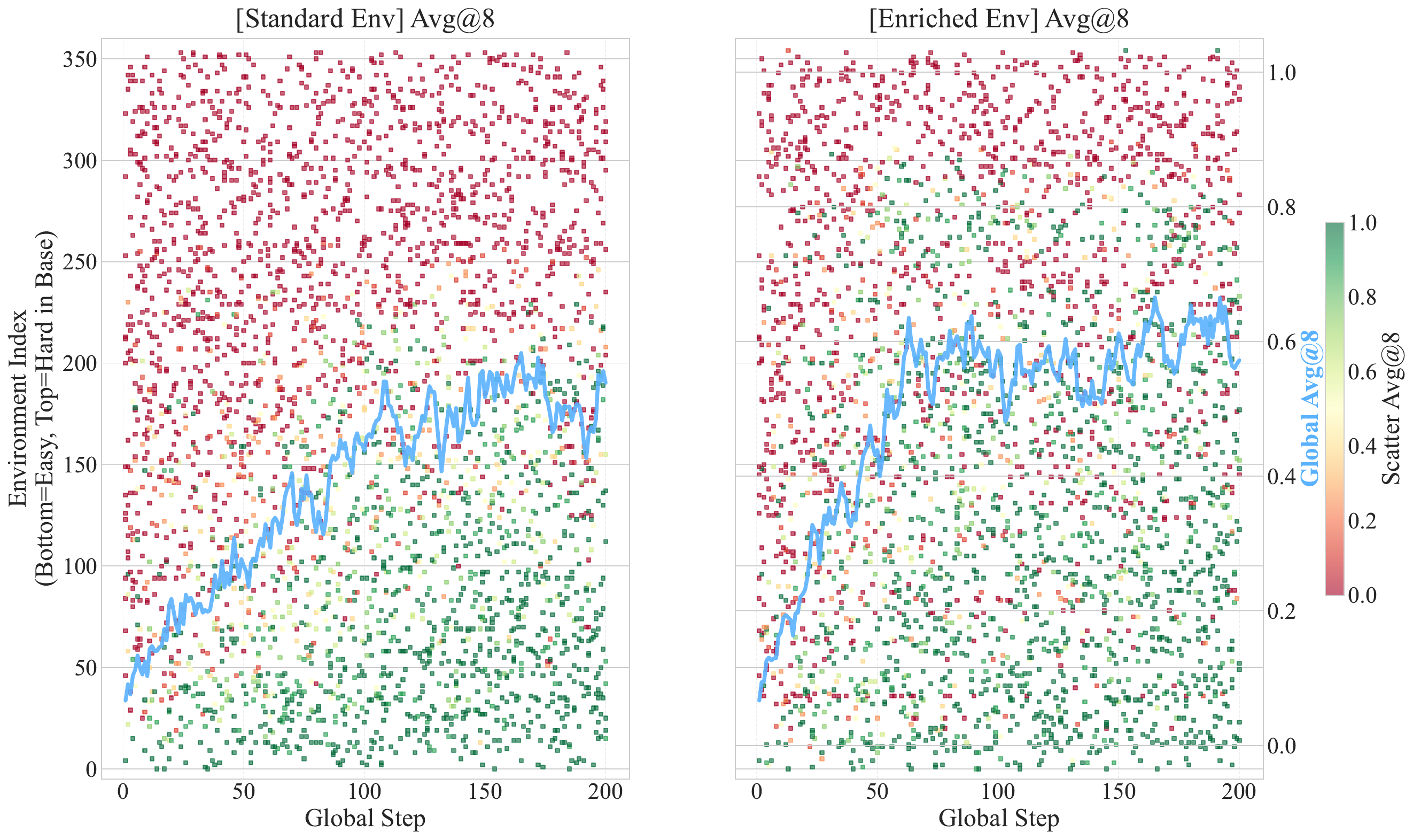}
                    \end{minipage}
    \hfill
        \begin{minipage}{0.33\linewidth}
        \centering
        \includegraphics[width=\linewidth]{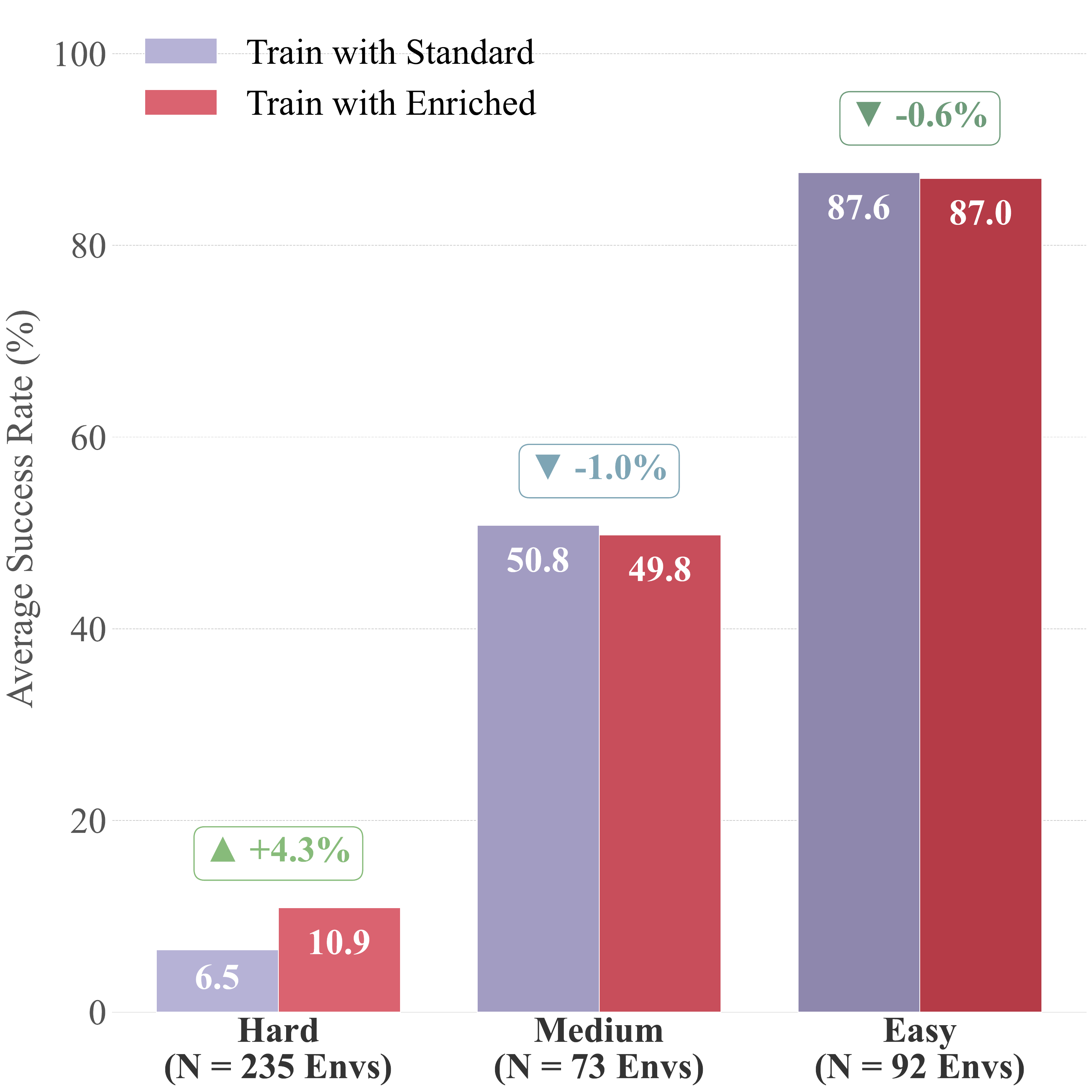}
                    \end{minipage}
    \caption{
\textbf{Left \& Middle:} Exploration dynamics of Qwen3-4B on BFCL under standard environments and FEEs. The x-axis denotes training steps and the y-axis denotes environment indices ordered by difficulty (bottom: easy, top: hard). Each point represents the avg@8 success rate of a sampled environment, while the blue curve shows the global average avg@8. 
\textbf{Right:} Validation success rates of models trained in standard environments and FEEs, evaluated on standard environments partitioned into hard, medium, and easy difficulty tiers.
}
        \label{fig:exploration_exp}
\end{figure*}

\textbf{Exp 2.}
We examine whether FEEs facilitate more effective and persistent state-space exploration from both the training and evaluation perspectives. Specifically, during training, we monitor Qwen3-4B on the BFCL benchmark by calculating the average success rate across 8 rollouts for each sampled environment. To assess the persistence of these capabilities during evaluation, we partition 400 environments into easy ($[0.66, 1]$), medium ($[0.33,0.66]$), and hard ($[0, 0.33]$) difficulty tiers based on the success rates achieved by models trained in standard environments, and then evaluate models trained in FEEs across these tiers.

\textbf{Res 2.} 
The experimental results are illustrated in Figure \ref{fig:exploration_exp}, yielding the following analysis. (1) From the training perspective, FEEs improve the agent's exploration efficiency during RL training. For instance, the FEE-trained agent exhibits a significantly higher density of green squares in the later stages of training,  whereas the standard baseline remains dominated by red squares across many environments. This suggests that enriched feedback enables the agent to explore and master a broader range of environment states. (2) From the evaluation perspective, the exploration capability learned with FEEs transfers effectively to standard environments. For instance, the FEE-trained model consistently outperforms the baseline across all difficulty levels, achieving a notable $4.3\%$ improvement on ``hard'' environments. This indicates that the agent develops a more proactive exploration strategy that allows it to navigate low-probability states even without enriched feedback.

\textbf{Takeaway 2.}  FEEs encourage persistent exploration that is internalized by the policy and remains effective in standard, high-difficulty environments.

\subsection{Feedback Internalization}

\textbf{Exp 3.}
We investigate whether the enriched information in FEEs is internalized into the agent’s policy weights or merely functions as a temporary inference-time hint. To study this, we design a controlled prediction experiment based on the GorillaFileSystem environment in BFCL. Specifically, we select a subset of enriched file-system operations, including \texttt{cd}, \texttt{mkdir}, \texttt{touch}, \texttt{rm}, \texttt{rmdir}, \texttt{mv}, \texttt{cp}, and \texttt{echo}. While standard feedback only provides standard tool outputs, the enriched feedback additionally includes the current absolute path. We then formulate multiple-choice probe questions that asks the model to predict the expected tool output after executing a command, and compare the output probabilities assigned to the original and enriched feedback options. Details of these probe questions can be found in Appendix \ref{appendix:internalization_exp}. We further apply these probe questions to Qwen3-4B throughout the later stages of BFCL training, evaluating them every 10 training steps to monitor the evolution of knowledge internalization.

\textbf{Res 3.}
Figure \ref{fig:internalization_exp} illustrates the probability margin between the original and enriched feedback options across training steps. When tested in standard environments, models trained with FEEs assign progressively higher probabilities to the enriched feedback option, whereas standard-trained models remain consistently biased toward the original tool output with negligible variation.

\textbf{Takeaway 3.}  
FEE-trained agents internalize the enriched information rather than treating it as a temporary inference-time hint.

\setcounter{figure}{5}

\Needspace{23\baselineskip}
\subsection{Intra-group Feedback Consistency}

\begin{wrapfigure}{r}{0.48\textwidth}
    \centering
    \includegraphics[width=\linewidth]{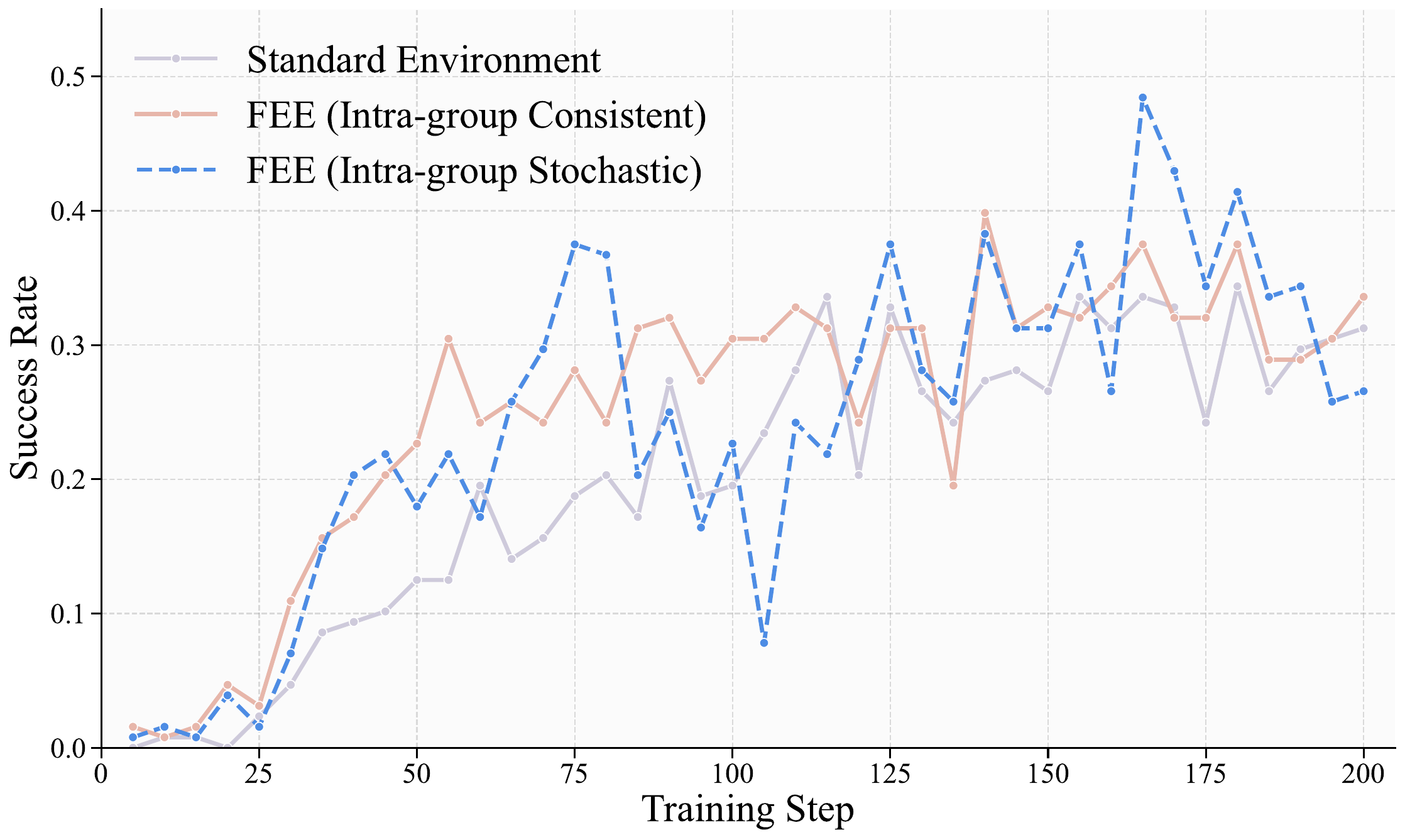}
    \caption{Validation success rates of Qwen3-4B in standard SciWorld environments during GRPO training under different feedback consistency settings.}
    \label{fig:consistency_exp}
\end{wrapfigure}

\textbf{Exp 4.}
We investigate whether the enriched environmental feedback should remain consistent or stochastic within a single rollout group, as feedback enrichment is typically injected with a probability of 0.5 during training. To study this, we compare two training configurations: a stochastic setting where different enriched feedback variants are randomly assigned within the same sampling group, and a consistent setting where all agents in a group receive identical feedback for a given state. Specifically, we conduct this experiment by training Qwen3-4B on SciWorld with GRPO for 200 training steps and monitor the validation performance in standard environments.

\textbf{Res 4.}
The results show that intra-group feedback consistency is crucial for stable optimization. As illustrated in Figure \ref{fig:consistency_exp}, the configuration with consistent intra-group feedback exhibits a steady increase in success rate throughout training. In contrast, providing diverse feedback within a single group leads to severe instability, characterized by erratic performance fluctuations with sharp plunges and sudden spikes. This may stem from the nature of group-based agentic RL, where advantages are estimated within each sampling group. Excessive stochasticity in intra-group feedback can distort advantage estimation, leading to noisy optimization signals and unstable convergence.

\textbf{Takeaway 4.}  
Intra-group feedback consistency is a prerequisite for stable optimization, as excessive diversity within a single rollout group triggers severe performance volatility.

\section{Related Work}

\subsection{RL for Long-horizon Tasks}

Agentic language models tackle long-horizon tasks by integrating natural language reasoning with grounded actions such as tool manipulation \citep{yao2024tau-benchBenchmarkTool-agent-user,wang2025mcp-benchBenchmarkingTool-using,trivedi2024appworldControllableWorld,lei2025mcpverseExpansiveReal-world}, web navigation \citep{yao2022webshopTowardsScalable,zhou2024webarenaRealisticWeb,koh2024visualwebarenaEvaluatingMultimodal,rawles2025androidworldDynamicBenchmarking,xie2024osworldBenchmarkingMultimodal}, and code execution \citep{ouyang2025kernelbenchCanLlms,merrill2026terminal-benchBenchmarkingAgents,yang2025swe-benchMultimodalDo}, requiring an extended sequence of steps to achieve an ultimate objective. 
To enhance the capabilities of LLMs serving as the backbone of agentic systems, reinforcement-learning-based approaches have been instrumental \citep{zhang2025rlvmr,wang2025ragenUnderstandingSelf-evolution,feng2025group-in-groupPolicyOptimization,jin2025search-r1TrainingLlms,hu2026seeupo}. Typically, a supervised fine-tuning warm-up phase is conducted prior to reinforcement learning to boost the agent's initial capabilities \citep{li2025websailor,wei2025webagent-r1,lu2025ui-s1,jin2025videomemEnhancingUltra-long}. However, in this paper, we introduce a complementary environment-side enrichment framework that is orthogonal to existing agent-centric optimization algorithms and refinement techniques.

\subsection{Environments for Evolving Agents}

As modern LLMs evolve into autonomous agents capable of sequential decision making in complex environments, training these agents within interactive, gym-style environments is becoming a standard practice \citep{environment_survey,liu2025gemGymAgentic,stojanovski2025reasoningGymReasoning,meng2026gymvunifiedvisionenvironment,aggarwal2026gymanythingturnsoftwareagent}. Consequently, the focus of research has evolved from scaling static datasets \citep{wang2023self-instruct,xu2024wizardlm} to scaling the complexity and diversity of interactive environments \citep{fang2025towardsGeneralAgentic,song2026envscalerScalingTool-interactive,gandhi2026endlessTerminals,tu2026scaleenvScalingEnvironment}. In the era of scaling data, to solve the reward sparsity problem, many work focuses on on modulating task difficulty by incorporating linguistic hints as a form of scaffolding to facilitate model training 
\citep{li2025questaExpandingReasoning,zhang2025scaf-grpoScaffoldedGroup,zhang2025adhintAdaptiveHints,zhang2025stephintMulti-levelStepwise,huang2025boostingMllmReasoning}.  However, in the era of scaling enironments, how to  systematically reconstruct multi-turn, dynamic environments to facilitate agent evolution remains relatively under-explored.

\section{Conclusion}

Training long-horizon LLM agents with RL remains challenging due to sparse rewards and ineffective exploration. In this work, we shift the focus from agent-side warming to environment-side adaptation, and propose a general strategy for constructing effective feedback-enriched environments. By systematically designing what feedback to provide and when to deliver it, the resulting FEEs consistently improve RL training across multiple benchmarks, model scales, and optimization algorithms. Beyond performance gains, our findings show that FEEs stabilize training dynamics, encourage proactive state-space exploration, internalize environmental guidance into policy weights, and highlight intra-group feedback consistency as an important condition for stable optimization.

\section*{Limitations}

Despite the promising effectiveness of FEEs, our study still has several limitations. First, constructing feedback-enriched environments requires environment-specific design choices and hyperparameters. For example, stage-dependent settings such as the definitions of early and late phases in intra-episode exploration and inter-episode evolution are manually specified in our experiments, while their sensitivity and optimal configurations remain underexplored. Second, although we validate FEEs on SciWorld and BFCL, we do not conduct broader evaluations across a wider range of agent benchmarks, leaving the generalization ability of our strategy insufficiently studied. Third, we observe clear limitations of our approach in more challenging environments. In particular, when training Qwen3-4B and 8B on AppWorld \citep{trivedi2024appworldControllableWorld}, introducing enriched feedback still fails to produce positive rewards, with training remaining trapped in zero-reward trajectories. This suggests that the effectiveness of FEEs is not universal and that richer environment adaptation strategies for extremely sparse long-horizon settings warrant further investigation.

\section*{Ethics and Artifact Use Statement}

\textbf{Potential risks.}
We do not identify significant potential risks associated with this work. Our study focuses on improving reinforcement learning training for LLM agents in benchmarked long-horizon environments through environment-side feedback design. The proposed method neither introduces deployment-facing systems nor involves sensitive data, human subjects, safety-critical decision making, or high-risk real-world applications. The feedback-enriched environments are constructed within controlled research benchmarks and are intended solely for studying training dynamics and agent learning behaviors.

\textbf{Artifacts, licenses, and intended use.}
Our work uses official open-source codebases and benchmark environments released by prior work. All utilized artifacts are appropriately cited in the paper. Documentation, implementation details, and usage instructions for these artifacts are publicly available at their corresponding official repositories and project websites.

\textbf{Data privacy and content safety.}
Our study uses only publicly available benchmark tasks and open-source research environments, without involving personal data, sensitive content, or human participant data.

\textbf{Use of AI assistants.}
AI assistants were used solely for minor writing refinement and language polishing.

\FloatBarrier
\clearpage
\bibliographystyle{latex/acl_natbib}
\bibliography{latex/custom}

@article{chen2025reinforcementLearningLong-horizon,
  author       = {Kevin Chen and
                  Marco F. Cusumano{-}Towner and
                  Brody Huval and
                  Aleksei Petrenko and
                  Jackson Hamburger and
                  Vladlen Koltun and
                  Philipp Kr{\"{a}}henb{\"{u}}hl},
  title        = {Reinforcement Learning for Long-Horizon Interactive {LLM} Agents},
  journal      = {CoRR},
  volume       = {abs/2502.01600},
  year         = {2025},
  url          = {https://doi.org/10.48550/arXiv.2502.01600},
  doi          = {10.48550/ARXIV.2502.01600},
  eprinttype    = {arXiv},
  eprint       = {2502.01600},
  bibsource    = {dblp computer science bibliography, https://dblp.org}
}

@article{shao2024deepseekmathPushingLimits,
  author       = {Zhihong Shao and
                  Peiyi Wang and
                  Qihao Zhu and
                  Runxin Xu and
                  Junxiao Song and
                  Mingchuan Zhang and
                  Y. K. Li and
                  Y. Wu and
                  Daya Guo},
  title        = {DeepSeekMath: Pushing the Limits of Mathematical Reasoning in Open
                  Language Models},
  journal      = {CoRR},
  volume       = {abs/2402.03300},
  year         = {2024},
  url          = {https://doi.org/10.48550/arXiv.2402.03300},
  doi          = {10.48550/ARXIV.2402.03300},
  eprinttype    = {arXiv},
  eprint       = {2402.03300},
  bibsource    = {dblp computer science bibliography, https://dblp.org}
}

@article{yu2025dapoOpen-sourceLlm,
  author       = {Qiying Yu and
                  Zheng Zhang and
                  Ruofei Zhu and
                  Yufeng Yuan and
                  Xiaochen Zuo and
                  Yu Yue and
                  Tiantian Fan and
                  Gaohong Liu and
                  Lingjun Liu and
                  Xin Liu and
                  Haibin Lin and
                  Zhiqi Lin and
                  Bole Ma and
                  Guangming Sheng and
                  Yuxuan Tong and
                  Chi Zhang and
                  Mofan Zhang and
                  Wang Zhang and
                  Hang Zhu and
                  Jinhua Zhu and
                  Jiaze Chen and
                  Jiangjie Chen and
                  Chengyi Wang and
                  Hongli Yu and
                  Weinan Dai and
                  Yuxuan Song and
                  Xiangpeng Wei and
                  Hao Zhou and
                  Jingjing Liu and
                  Wei{-}Ying Ma and
                  Ya{-}Qin Zhang and
                  Lin Yan and
                  Mu Qiao and
                  Yonghui Wu and
                  Mingxuan Wang},
  title        = {{DAPO:} An Open-Source {LLM} Reinforcement Learning System at Scale},
  journal      = {CoRR},
  volume       = {abs/2503.14476},
  year         = {2025},
  url          = {https://doi.org/10.48550/arXiv.2503.14476},
  doi          = {10.48550/ARXIV.2503.14476},
  eprinttype    = {arXiv},
  eprint       = {2503.14476},
  bibsource    = {dblp computer science bibliography, https://dblp.org}
}

@article{zheng2025groupSequencePolicy,
  author       = {Chujie Zheng and
                  Shixuan Liu and
                  Mingze Li and
                  Xiong{-}Hui Chen and
                  Bowen Yu and
                  Chang Gao and
                  Kai Dang and
                  Yuqiong Liu and
                  Rui Men and
                  An Yang and
                  Jingren Zhou and
                  Junyang Lin},
  title        = {Group Sequence Policy Optimization},
  journal      = {CoRR},
  volume       = {abs/2507.18071},
  year         = {2025},
  url          = {https://doi.org/10.48550/arXiv.2507.18071},
  doi          = {10.48550/ARXIV.2507.18071},
  eprinttype    = {arXiv},
  eprint       = {2507.18071},
  bibsource    = {dblp computer science bibliography, https://dblp.org}
}

@article{wang2025harnessingUncertainty,
  author       = {Jiawei Wang and
                  Jiacai Liu and
                  Yuqian Fu and
                  Yingru Li and
                  Xintao Wang and
                  Yuan Lin and
                  Yu Yue and
                  Lin Zhang and
                  Yang Wang and
                  Ke Wang},
  title        = {Harnessing Uncertainty: Entropy-Modulated Policy Gradients for Long-Horizon
                  {LLM} Agents},
  journal      = {CoRR},
  volume       = {abs/2509.09265},
  year         = {2025},
  url          = {https://doi.org/10.48550/arXiv.2509.09265},
  doi          = {10.48550/ARXIV.2509.09265},
  eprinttype    = {arXiv},
  eprint       = {2509.09265},
  bibsource    = {dblp computer science bibliography, https://dblp.org}
}

@inproceedings{wang2022scienceworld,
  author       = {Ruoyao Wang and
                  Peter A. Jansen and
                  Marc{-}Alexandre C{\^{o}}t{\'{e}} and
                  Prithviraj Ammanabrolu},
  editor       = {Yoav Goldberg and
                  Zornitsa Kozareva and
                  Yue Zhang},
  title        = {ScienceWorld: Is your Agent Smarter than a 5th Grader?},
  booktitle    = {Proceedings of the 2022 Conference on Empirical Methods in Natural
                  Language Processing, {EMNLP} 2022, Abu Dhabi, United Arab Emirates,
                  December 7-11, 2022},
  pages        = {11279--11298},
  publisher    = {Association for Computational Linguistics},
  year         = {2022},
  url          = {https://doi.org/10.18653/v1/2022.emnlp-main.775},
  doi          = {10.18653/V1/2022.EMNLP-MAIN.775},
  bibsource    = {dblp computer science bibliography, https://dblp.org}
}

@inproceedings{patil2025berkeleyFunctionCallingLeaderboard,
  author       = {Shishir G. Patil and
                  Huanzhi Mao and
                  Fanjia Yan and
                  Charlie Cheng{-}Jie Ji and
                  Vishnu Suresh and
                  Ion Stoica and
                  Joseph E. Gonzalez},
  editor       = {Aarti Singh and
                  Maryam Fazel and
                  Daniel Hsu and
                  Simon Lacoste{-}Julien and
                  Felix Berkenkamp and
                  Tegan Maharaj and
                  Kiri Wagstaff and
                  Jerry Zhu},
  title        = {The Berkeley Function Calling Leaderboard {(BFCL):} From Tool Use
                  to Agentic Evaluation of Large Language Models},
  booktitle    = {Forty-second International Conference on Machine Learning, {ICML}
                  2025, Vancouver, BC, Canada, July 13-19, 2025},
  series       = {Proceedings of Machine Learning Research},
  volume       = {267},
  publisher    = {{PMLR} / OpenReview.net},
  year         = {2025},
  url          = {https://proceedings.mlr.press/v267/patil25a.html},
  bibsource    = {dblp computer science bibliography, https://dblp.org}
}

@article{glazer2024frontiermath,
  author       = {Elliot Glazer and
                  Ege Erdil and
                  Tamay Besiroglu and
                  Diego Chicharro and
                  Evan Chen and
                  Alex Gunning and
                  Caroline Falkman Olsson and
                  Jean{-}Stanislas Denain and
                  Anson Ho and
                  Emily de Oliveira Santos and
                  Olli J{\"{a}}rviniemi and
                  Matthew Barnett and
                  Robert Sandler and
                  Matej Vrzala and
                  Jaime Sevilla and
                  Qiuyu Ren and
                  Elizabeth Pratt and
                  Lionel Levine and
                  Grant Barkley and
                  Natalie Stewart and
                  Bogdan Grechuk and
                  Tetiana Grechuk and
                  Shreepranav Varma Enugandla and
                  Mark Wildon},
  title        = {FrontierMath: {A} Benchmark for Evaluating Advanced Mathematical Reasoning
                  in {AI}},
  journal      = {CoRR},
  volume       = {abs/2411.04872},
  year         = {2024},
  url          = {https://doi.org/10.48550/arXiv.2411.04872},
  doi          = {10.48550/ARXIV.2411.04872},
  eprinttype    = {arXiv},
  eprint       = {2411.04872},
  bibsource    = {dblp computer science bibliography, https://dblp.org}
}

@article{du2025nemotron-math,
  author       = {Wei Du and
                  Shubham Toshniwal and
                  Branislav Kisacanin and
                  Sadegh Mahdavi and
                  Ivan Moshkov and
                  George Armstrong and
                  Stephen Ge and
                  Edgar Minasyan and
                  Feng Chen and
                  Igor Gitman},
  title        = {Nemotron-Math: Efficient Long-Context Distillation of Mathematical
                  Reasoning from Multi-Mode Supervision},
  journal      = {CoRR},
  volume       = {abs/2512.15489},
  year         = {2025},
  url          = {https://doi.org/10.48550/arXiv.2512.15489},
  doi          = {10.48550/ARXIV.2512.15489},
  eprinttype    = {arXiv},
  eprint       = {2512.15489},
  bibsource    = {dblp computer science bibliography, https://dblp.org}
}

@article{merrill2026terminal-bench,
  author       = {Mike A. Merrill and
                  Alexander Glenn Shaw and
                  Nicholas Carlini and
                  Boxuan Li and
                  Harsh Raj and
                  Ivan Bercovich and
                  Lin Shi and
                  Jeong Yeon Shin and
                  Thomas Walshe and
                  Estefany Kelly Buchanan and
                  Junhong Shen and
                  Guanghao Ye and
                  Haowei Lin and
                  Jason Poulos and
                  Maoyu Wang and
                  Marianna Nezhurina and
                  Jenia Jitsev and
                  Di Lu and
                  Orfeas Menis{-}Mastromichalakis and
                  Zhiwei Xu and
                  Zizhao Chen and
                  Yue Liu and
                  Robert Zhang and
                  Leon Liangyu Chen and
                  Anurag Kashyap and
                  Jan{-}Lucas Uslu and
                  Jeffrey Li and
                  Jianbo Wu and
                  Minghao Yan and
                  Song Bian and
                  Vedang Sharma and
                  Ke Sun and
                  Steven Dillmann and
                  Akshay Anand and
                  Andrew Lanpouthakoun and
                  Bardia Koopah and
                  Changran Hu and
                  Etash Kumar Guha and
                  Gabriel H. S. Dreiman and
                  Jiacheng Zhu and
                  Karl Krauth and
                  Li Zhong and
                  Niklas Muennighoff and
                  Robert Amanfu and
                  Shangyin Tan and
                  Shreyas Pimpalgaonkar and
                  Tushar Aggarwal and
                  Xiangning Lin and
                  Xin Lan and
                  Xuandong Zhao and
                  Yiqing Liang and
                  Yuanli Wang and
                  Zilong Wang and
                  Changzhi Zhou and
                  David Heineman and
                  Hange Liu and
                  Harsh Trivedi and
                  John Yang and
                  Junhong Lin and
                  Manish Shetty and
                  Michael Yang and
                  Nabil Omi and
                  Negin Raoof and
                  Shanda Li and
                  Terry Yue Zhuo and
                  Wuwei Lin and
                  Yiwei Dai and
                  Yuxin Wang and
                  Wenhao Chai and
                  Shang Zhou and
                  Dariush Wahdany and
                  Ziyu She and
                  Jiaming Hu and
                  Zhikang Dong and
                  Yuxuan Zhu and
                  Sasha Cui and
                  Ahson Saiyed and
                  Arinbj{\"{o}}rn Kolbeinsson and
                  Jesse Hu and
                  Christopher Michael Rytting and
                  Ryan Marten and
                  Yixin Wang and
                  Alex Dimakis and
                  Andy Konwinski and
                  Ludwig Schmidt},
  title        = {Terminal-Bench: Benchmarking Agents on Hard, Realistic Tasks in Command
                  Line Interfaces},
  journal      = {CoRR},
  volume       = {abs/2601.11868},
  year         = {2026},
  url          = {https://doi.org/10.48550/arXiv.2601.11868},
  doi          = {10.48550/ARXIV.2601.11868},
  eprinttype    = {arXiv},
  eprint       = {2601.11868},
  bibsource    = {dblp computer science bibliography, https://dblp.org}
}

@inproceedings{jimenez2024swe-bench,
  author       = {Carlos E. Jimenez and
                  John Yang and
                  Alexander Wettig and
                  Shunyu Yao and
                  Kexin Pei and
                  Ofir Press and
                  Karthik R. Narasimhan},
  title        = {SWE-bench: Can Language Models Resolve Real-world Github Issues?},
  booktitle    = {The Twelfth International Conference on Learning Representations,
                  {ICLR} 2024, Vienna, Austria, May 7-11, 2024},
  publisher    = {OpenReview.net},
  year         = {2024},
  url          = {https://openreview.net/forum?id=VTF8yNQM66},
  bibsource    = {dblp computer science bibliography, https://dblp.org}
}

@article{gpt4,
  author       = {OpenAI},
  title        = {{GPT-4} Technical Report},
  journal      = {CoRR},
  volume       = {abs/2303.08774},
  year         = {2023},
  url          = {https://doi.org/10.48550/arXiv.2303.08774},
  doi          = {10.48550/ARXIV.2303.08774},
  eprinttype    = {arXiv},
  eprint       = {2303.08774},
  bibsource    = {dblp computer science bibliography, https://dblp.org}
}

@article{qwen3,
  author       = {An Yang and
                  Anfeng Li and
                  Baosong Yang and
                  Beichen Zhang and
                  Binyuan Hui and
                  Bo Zheng and
                  Bowen Yu and
                  Chang Gao and
                  Chengen Huang and
                  Chenxu Lv and
                  Chujie Zheng and
                  Dayiheng Liu and
                  Fan Zhou and
                  Fei Huang and
                  Feng Hu and
                  Hao Ge and
                  Haoran Wei and
                  Huan Lin and
                  Jialong Tang and
                  Jian Yang and
                  Jianhong Tu and
                  Jianwei Zhang and
                  Jian Yang and
                  Jiaxi Yang and
                  Jingren Zhou and
                  Junyang Lin and
                  Kai Dang and
                  Keqin Bao and
                  Kexin Yang and
                  Le Yu and
                  Lianghao Deng and
                  Mei Li and
                  Mingfeng Xue and
                  Mingze Li and
                  Pei Zhang and
                  Peng Wang and
                  Qin Zhu and
                  Rui Men and
                  Ruize Gao and
                  Shixuan Liu and
                  Shuang Luo and
                  Tianhao Li and
                  Tianyi Tang and
                  Wenbiao Yin and
                  Xingzhang Ren and
                  Xinyu Wang and
                  Xinyu Zhang and
                  Xuancheng Ren and
                  Yang Fan and
                  Yang Su and
                  Yichang Zhang and
                  Yinger Zhang and
                  Yu Wan and
                  Yuqiong Liu and
                  Zekun Wang and
                  Zeyu Cui and
                  Zhenru Zhang and
                  Zhipeng Zhou and
                  Zihan Qiu},
  title        = {Qwen3 Technical Report},
  journal      = {CoRR},
  volume       = {abs/2505.09388},
  year         = {2025},
  url          = {https://doi.org/10.48550/arXiv.2505.09388},
  doi          = {10.48550/ARXIV.2505.09388},
  eprinttype    = {arXiv},
  eprint       = {2505.09388},
  bibsource    = {dblp computer science bibliography, https://dblp.org}
}

@article{llama3,
  author       = {Abhimanyu Dubey and
                  Abhinav Jauhri and
                  Abhinav Pandey and
                  Abhishek Kadian and
                  Ahmad Al{-}Dahle and
                  Aiesha Letman and
                  Akhil Mathur and
                  Alan Schelten and
                  Amy Yang and
                  Angela Fan and
                  Anirudh Goyal and
                  Anthony Hartshorn and
                  Aobo Yang and
                  Archi Mitra and
                  Archie Sravankumar and
                  Artem Korenev and
                  Arthur Hinsvark and
                  Arun Rao and
                  Aston Zhang and
                  Aur{\'{e}}lien Rodriguez and
                  Austen Gregerson and
                  Ava Spataru and
                  Baptiste Rozi{\`{e}}re and
                  Bethany Biron and
                  Binh Tang and
                  Bobbie Chern and
                  Charlotte Caucheteux and
                  Chaya Nayak and
                  Chloe Bi and
                  Chris Marra and
                  Chris McConnell and
                  Christian Keller and
                  Christophe Touret and
                  Chunyang Wu and
                  Corinne Wong and
                  Cristian Canton Ferrer and
                  Cyrus Nikolaidis and
                  Damien Allonsius and
                  Daniel Song and
                  Danielle Pintz and
                  Danny Livshits and
                  David Esiobu and
                  Dhruv Choudhary and
                  Dhruv Mahajan and
                  Diego Garcia{-}Olano and
                  Diego Perino and
                  Dieuwke Hupkes and
                  Egor Lakomkin and
                  Ehab AlBadawy and
                  Elina Lobanova and
                  Emily Dinan and
                  Eric Michael Smith and
                  Filip Radenovic and
                  Frank Zhang and
                  Gabriel Synnaeve and
                  Gabrielle Lee and
                  Georgia Lewis Anderson and
                  Graeme Nail and
                  Gr{\'{e}}goire Mialon and
                  Guan Pang and
                  Guillem Cucurell and
                  Hailey Nguyen and
                  Hannah Korevaar and
                  Hu Xu and
                  Hugo Touvron and
                  Iliyan Zarov and
                  Imanol Arrieta Ibarra and
                  Isabel M. Kloumann and
                  Ishan Misra and
                  Ivan Evtimov and
                  Jade Copet and
                  Jaewon Lee and
                  Jan Geffert and
                  Jana Vranes and
                  Jason Park and
                  Jay Mahadeokar and
                  Jeet Shah and
                  Jelmer van der Linde and
                  Jennifer Billock and
                  Jenny Hong and
                  Jenya Lee and
                  Jeremy Fu and
                  Jianfeng Chi and
                  Jianyu Huang and
                  Jiawen Liu and
                  Jie Wang and
                  Jiecao Yu and
                  Joanna Bitton and
                  Joe Spisak and
                  Jongsoo Park and
                  Joseph Rocca and
                  Joshua Johnstun and
                  Joshua Saxe and
                  Junteng Jia and
                  Kalyan Vasuden Alwala and
                  Kartikeya Upasani and
                  Kate Plawiak and
                  Ke Li and
                  Kenneth Heafield and
                  Kevin Stone and
                  et al.},
  title        = {The Llama 3 Herd of Models},
  journal      = {CoRR},
  volume       = {abs/2407.21783},
  year         = {2024},
  url          = {https://doi.org/10.48550/arXiv.2407.21783},
  doi          = {10.48550/ARXIV.2407.21783},
  eprinttype    = {arXiv},
  eprint       = {2407.21783},
  bibsource    = {dblp computer science bibliography, https://dblp.org}
}

@article{deepseekr1,
  author       = {DeepSeek{-}AI and
                  Daya Guo and
                  Dejian Yang and
                  Haowei Zhang and
                  Junxiao Song and
                  Ruoyu Zhang and
                  Runxin Xu and
                  Qihao Zhu and
                  Shirong Ma and
                  Peiyi Wang and
                  Xiao Bi and
                  Xiaokang Zhang and
                  Xingkai Yu and
                  Yu Wu and
                  Z. F. Wu and
                  Zhibin Gou and
                  Zhihong Shao and
                  Zhuoshu Li and
                  Ziyi Gao and
                  Aixin Liu and
                  Bing Xue and
                  Bingxuan Wang and
                  Bochao Wu and
                  Bei Feng and
                  Chengda Lu and
                  Chenggang Zhao and
                  Chengqi Deng and
                  Chenyu Zhang and
                  Chong Ruan and
                  Damai Dai and
                  Deli Chen and
                  Dongjie Ji and
                  Erhang Li and
                  Fangyun Lin and
                  Fucong Dai and
                  Fuli Luo and
                  Guangbo Hao and
                  Guanting Chen and
                  Guowei Li and
                  H. Zhang and
                  Han Bao and
                  Hanwei Xu and
                  Haocheng Wang and
                  Honghui Ding and
                  Huajian Xin and
                  Huazuo Gao and
                  Hui Qu and
                  Hui Li and
                  Jianzhong Guo and
                  Jiashi Li and
                  Jiawei Wang and
                  Jingchang Chen and
                  Jingyang Yuan and
                  Junjie Qiu and
                  Junlong Li and
                  J. L. Cai and
                  Jiaqi Ni and
                  Jian Liang and
                  Jin Chen and
                  Kai Dong and
                  Kai Hu and
                  Kaige Gao and
                  Kang Guan and
                  Kexin Huang and
                  Kuai Yu and
                  Lean Wang and
                  Lecong Zhang and
                  Liang Zhao and
                  Litong Wang and
                  Liyue Zhang and
                  Lei Xu and
                  Leyi Xia and
                  Mingchuan Zhang and
                  Minghua Zhang and
                  Minghui Tang and
                  Meng Li and
                  Miaojun Wang and
                  Mingming Li and
                  Ning Tian and
                  Panpan Huang and
                  Peng Zhang and
                  Qiancheng Wang and
                  Qinyu Chen and
                  Qiushi Du and
                  Ruiqi Ge and
                  Ruisong Zhang and
                  Ruizhe Pan and
                  Runji Wang and
                  R. J. Chen and
                  R. L. Jin and
                  Ruyi Chen and
                  Shanghao Lu and
                  Shangyan Zhou and
                  Shanhuang Chen and
                  Shengfeng Ye and
                  Shiyu Wang and
                  Shuiping Yu and
                  Shunfeng Zhou and
                  Shuting Pan and
                  S. S. Li},
  title        = {DeepSeek-R1: Incentivizing Reasoning Capability in LLMs via Reinforcement
                  Learning},
  journal      = {CoRR},
  volume       = {abs/2501.12948},
  year         = {2025},
  url          = {https://doi.org/10.48550/arXiv.2501.12948},
  doi          = {10.48550/ARXIV.2501.12948},
  eprinttype    = {arXiv},
  eprint       = {2501.12948},
  bibsource    = {dblp computer science bibliography, https://dblp.org}
}

@inproceedings{zhou2024webarena,
  author       = {Shuyan Zhou and
                  Frank F. Xu and
                  Hao Zhu and
                  Xuhui Zhou and
                  Robert Lo and
                  Abishek Sridhar and
                  Xianyi Cheng and
                  Tianyue Ou and
                  Yonatan Bisk and
                  Daniel Fried and
                  Uri Alon and
                  Graham Neubig},
  title        = {WebArena: {A} Realistic Web Environment for Building Autonomous Agents},
  booktitle    = {The Twelfth International Conference on Learning Representations,
                  {ICLR} 2024, Vienna, Austria, May 7-11, 2024},
  publisher    = {OpenReview.net},
  year         = {2024},
  url          = {https://openreview.net/forum?id=oKn9c6ytLx},
  bibsource    = {dblp computer science bibliography, https://dblp.org}
}

@article{bai2026webgym,
  author       = {Hao Bai and
                  Alexey Taymanov and
                  Tong Zhang and
                  Aviral Kumar and
                  Spencer Whitehead},
  title        = {WebGym: Scaling Training Environments for Visual Web Agents with Realistic
                  Tasks},
  journal      = {CoRR},
  volume       = {abs/2601.02439},
  year         = {2026},
  url          = {https://doi.org/10.48550/arXiv.2601.02439},
  doi          = {10.48550/ARXIV.2601.02439},
  eprinttype    = {arXiv},
  eprint       = {2601.02439},
  bibsource    = {dblp computer science bibliography, https://dblp.org}
}

@article{gandhi2026endlessTerminals,
  author       = {Kanishk Gandhi and
                  Shivam Garg and
                  Noah D. Goodman and
                  Dimitris Papailiopoulos},
  title        = {Endless Terminals: Scaling {RL} Environments for Terminal Agents},
  journal      = {CoRR},
  volume       = {abs/2601.16443},
  year         = {2026},
  url          = {https://doi.org/10.48550/arXiv.2601.16443},
  doi          = {10.48550/ARXIV.2601.16443},
  eprinttype    = {arXiv},
  eprint       = {2601.16443},
  bibsource    = {dblp computer science bibliography, https://dblp.org}
}

@inproceedings{yang2024swe-agent,
  author       = {John Yang and
                  Carlos E. Jimenez and
                  Alexander Wettig and
                  Kilian Lieret and
                  Shunyu Yao and
                  Karthik Narasimhan and
                  Ofir Press},
  editor       = {Amir Globersons and
                  Lester Mackey and
                  Danielle Belgrave and
                  Angela Fan and
                  Ulrich Paquet and
                  Jakub M. Tomczak and
                  Cheng Zhang},
  title        = {SWE-agent: Agent-Computer Interfaces Enable Automated Software Engineering},
  booktitle    = {Advances in Neural Information Processing Systems 38: Annual Conference
                  on Neural Information Processing Systems 2024, NeurIPS 2024, Vancouver,
                  BC, Canada, December 10 - 15, 2024},
  year         = {2024},
  url          = {http://papers.nips.cc/paper\_files/paper/2024/hash/5a7c947568c1b1328ccc5230172e1e7c-Abstract-Conference.html},
  bibsource    = {dblp computer science bibliography, https://dblp.org}
}

@article{zhang2025rlvmr,
  author       = {Zijing Zhang and
                  Ziyang Chen and
                  Mingxiao Li and
                  Zhaopeng Tu and
                  Xiaolong Li},
  title        = {{RLVMR:} Reinforcement Learning with Verifiable Meta-Reasoning Rewards
                  for Robust Long-Horizon Agents},
  journal      = {CoRR},
  volume       = {abs/2507.22844},
  year         = {2025},
  url          = {https://doi.org/10.48550/arXiv.2507.22844},
  doi          = {10.48550/ARXIV.2507.22844},
  eprinttype    = {arXiv},
  eprint       = {2507.22844},
  bibsource    = {dblp computer science bibliography, https://dblp.org}
}

@article{feng2025group-in-group,
  author       = {Lang Feng and
                  Zhenghai Xue and
                  Tingcong Liu and
                  Bo An},
  title        = {Group-in-Group Policy Optimization for {LLM} Agent Training},
  journal      = {CoRR},
  volume       = {abs/2505.10978},
  year         = {2025},
  url          = {https://doi.org/10.48550/arXiv.2505.10978},
  doi          = {10.48550/ARXIV.2505.10978},
  eprinttype    = {arXiv},
  eprint       = {2505.10978},
  bibsource    = {dblp computer science bibliography, https://dblp.org}
}

@article{chen2025scalingAgentLearning,
  author       = {Zhaorun Chen and
                  Zhuokai Zhao and
                  Kai Zhang and
                  Bo Liu and
                  Qi Qi and
                  Yifan Wu and
                  Tarun Kalluri and
                  Sara Cao and
                  Yuanhao Xiong and
                  Haibo Tong and
                  Huaxiu Yao and
                  Hengduo Li and
                  Jiacheng Zhu and
                  Xian Li and
                  Dawn Song and
                  Bo Li and
                  Jason Weston and
                  Dat Huynh},
  title        = {Scaling Agent Learning via Experience Synthesis},
  journal      = {CoRR},
  volume       = {abs/2511.03773},
  year         = {2025},
  url          = {https://doi.org/10.48550/arXiv.2511.03773},
  doi          = {10.48550/ARXIV.2511.03773},
  eprinttype    = {arXiv},
  eprint       = {2511.03773},
  bibsource    = {dblp computer science bibliography, https://dblp.org}
}

@article{00332025agentLearningEarly,
  author       = {Kai Zhang and
                  Xiangchao Chen and
                  Bo Liu and
                  Tianci Xue and
                  Zeyi Liao and
                  Zhihan Liu and
                  Xiyao Wang and
                  Yuting Ning and
                  Zhaorun Chen and
                  Xiaohan Fu and
                  Jian Xie and
                  Yuxuan Sun and
                  Boyu Gou and
                  Qi Qi and
                  Zihang Meng and
                  Jianwei Yang and
                  Ning Zhang and
                  Xian Li and
                  Ashish Shah and
                  Dat Huynh and
                  Hengduo Li and
                  Zi Yang and
                  Sara Cao and
                  Lawrence Jang and
                  Shuyan Zhou and
                  Jiacheng Zhu and
                  Huan Sun and
                  Jason Weston and
                  Yu Su and
                  Yifan Wu},
  title        = {Agent Learning via Early Experience},
  journal      = {CoRR},
  volume       = {abs/2510.08558},
  year         = {2025},
  url          = {https://doi.org/10.48550/arXiv.2510.08558},
  doi          = {10.48550/ARXIV.2510.08558},
  eprinttype    = {arXiv},
  eprint       = {2510.08558},
  bibsource    = {dblp computer science bibliography, https://dblp.org}
}

@article{kang2025quagmiresSft-rlPost-training,
  author       = {Feiyang Kang and
                  Michael Kuchnik and
                  Karthik Padthe and
                  Marin Vlastelica and
                  Ruoxi Jia and
                  Carole{-}Jean Wu and
                  Newsha Ardalani},
  title        = {Quagmires in {SFT-RL} Post-Training: When High {SFT} Scores Mislead
                  and What to Use Instead},
  journal      = {CoRR},
  volume       = {abs/2510.01624},
  year         = {2025},
  url          = {https://doi.org/10.48550/arXiv.2510.01624},
  doi          = {10.48550/ARXIV.2510.01624},
  eprinttype    = {arXiv},
  eprint       = {2510.01624},
  bibsource    = {dblp computer science bibliography, https://dblp.org}
}

@article{00012025acereason-nemotron11,
  author       = {Zihan Liu and
                  Zhuolin Yang and
                  Yang Chen and
                  Chankyu Lee and
                  Mohammad Shoeybi and
                  Bryan Catanzaro and
                  Wei Ping},
  title        = {AceReason-Nemotron 1.1: Advancing Math and Code Reasoning through
                  {SFT} and {RL} Synergy},
  journal      = {CoRR},
  volume       = {abs/2506.13284},
  year         = {2025},
  url          = {https://doi.org/10.48550/arXiv.2506.13284},
  doi          = {10.48550/ARXIV.2506.13284},
  eprinttype    = {arXiv},
  eprint       = {2506.13284},
  bibsource    = {dblp computer science bibliography, https://dblp.org}
}

@inproceedings{wen2025light-r1CurriculumSft,
  author       = {Liang Wen and
                  Yunke Cai and
                  Fenrui Xiao and
                  Xin He and
                  Qi An and
                  Zhenyu Duan and
                  Yimin Du and
                  Junchen Liu and
                  Lifu Tang and
                  Xiaowei Lv and
                  Haosheng Zou and
                  Yongchao Deng and
                  Shousheng Jia and
                  Xiangzheng Zhang},
  editor       = {Georg Rehm and
                  Yunyao Li},
  title        = {Light-R1: Curriculum SFT, {DPO} and {RL} for Long {COT} from Scratch
                  and Beyond},
  booktitle    = {Proceedings of the 63rd Annual Meeting of the Association for Computational
                  Linguistics (Volume 6: Industry Track), {ACL} 2025, Vienna, Austria,
                  July 27 - August 1, 2025},
  pages        = {318--327},
  publisher    = {Association for Computational Linguistics},
  year         = {2025},
  url          = {https://doi.org/10.18653/v1/2025.acl-industry.24},
  doi          = {10.18653/V1/2025.ACL-INDUSTRY.24},
  bibsource    = {dblp computer science bibliography, https://dblp.org}
}

@article{yao2024tau-benchBenchmarkTool-agent-user,
  author       = {Shunyu Yao and
                  Noah Shinn and
                  Pedram Razavi and
                  Karthik Narasimhan},
  title        = {{\(\tau\)}-bench: {A} Benchmark for Tool-Agent-User Interaction in
                  Real-World Domains},
  journal      = {CoRR},
  volume       = {abs/2406.12045},
  year         = {2024},
  url          = {https://doi.org/10.48550/arXiv.2406.12045},
  doi          = {10.48550/ARXIV.2406.12045},
  eprinttype   = {arXiv},
  eprint       = {2406.12045},
  bibsource    = {dblp computer science bibliography, https://dblp.org}
}

@article{wang2025mcp-benchBenchmarkingTool-using,
  author       = {Zhenting Wang and
                  Qi Chang and
                  Hemani Patel and
                  Shashank Biju and
                  Cheng{-}En Wu and
                  Quan Liu and
                  Aolin Ding and
                  Alireza Rezazadeh and
                  Ankit Shah and
                  Yujia Bao and
                  Eugene Siow},
  title        = {MCP-Bench: Benchmarking Tool-Using {LLM} Agents with Complex Real-World
                  Tasks via {MCP} Servers},
  journal      = {CoRR},
  volume       = {abs/2508.20453},
  year         = {2025},
  url          = {https://doi.org/10.48550/arXiv.2508.20453},
  doi          = {10.48550/ARXIV.2508.20453},
  eprinttype   = {arXiv},
  eprint       = {2508.20453},
  bibsource    = {dblp computer science bibliography, https://dblp.org}
}

@inproceedings{trivedi2024appworldControllableWorld,
  author       = {Harsh Trivedi and
                  Tushar Khot and
                  Mareike Hartmann and
                  Ruskin Manku and
                  Vinty Dong and
                  Edward Li and
                  Shashank Gupta and
                  Ashish Sabharwal and
                  Niranjan Balasubramanian},
  editor       = {Lun{-}Wei Ku and
                  Andre Martins and
                  Vivek Srikumar},
  title        = {AppWorld: {A} Controllable World of Apps and People for Benchmarking
                  Interactive Coding Agents},
  booktitle    = {Proceedings of the 62nd Annual Meeting of the Association for Computational
                  Linguistics (Volume 1: Long Papers), {ACL} 2024, Bangkok, Thailand,
                  August 11-16, 2024},
  pages        = {16022--16076},
  publisher    = {Association for Computational Linguistics},
  year         = {2024},
  url          = {https://doi.org/10.18653/v1/2024.acl-long.850},
  doi          = {10.18653/V1/2024.ACL-LONG.850},
  bibsource    = {dblp computer science bibliography, https://dblp.org}
}

@article{lei2025mcpverseExpansiveReal-world,
  author       = {Fei Lei and
                  Yibo Yang and
                  Wenxiu Sun and
                  Dahua Lin},
  title        = {MCPVerse: An Expansive, Real-World Benchmark for Agentic Tool Use},
  journal      = {CoRR},
  volume       = {abs/2508.16260},
  year         = {2025},
  url          = {https://doi.org/10.48550/arXiv.2508.16260},
  doi          = {10.48550/ARXIV.2508.16260},
  eprinttype   = {arXiv},
  eprint       = {2508.16260},
  bibsource    = {dblp computer science bibliography, https://dblp.org}
}

@inproceedings{yao2022webshopTowardsScalable,
  author       = {Shunyu Yao and
                  Howard Chen and
                  John Yang and
                  Karthik Narasimhan},
  editor       = {Sanmi Koyejo and
                  S. Mohamed and
                  A. Agarwal and
                  Danielle Belgrave and
                  K. Cho and
                  A. Oh},
  title        = {WebShop: Towards Scalable Real-World Web Interaction with Grounded
                  Language Agents},
  booktitle    = {Advances in Neural Information Processing Systems 35: Annual Conference
                  on Neural Information Processing Systems 2022, NeurIPS 2022, New Orleans,
                  LA, USA, November 28 - December 9, 2022},
  year         = {2022},
  url          = {http://papers.nips.cc/paper\_files/paper/2022/hash/82ad13ec01f9fe44c01cb91814fd7b8c-Abstract-Conference.html},
  bibsource    = {dblp computer science bibliography, https://dblp.org}
}

@inproceedings{zhou2024webarenaRealisticWeb,
  author       = {Shuyan Zhou and
                  Frank F. Xu and
                  Hao Zhu and
                  Xuhui Zhou and
                  Robert Lo and
                  Abishek Sridhar and
                  Xianyi Cheng and
                  Tianyue Ou and
                  Yonatan Bisk and
                  Daniel Fried and
                  Uri Alon and
                  Graham Neubig},
  title        = {WebArena: {A} Realistic Web Environment for Building Autonomous Agents},
  booktitle    = {The Twelfth International Conference on Learning Representations,
                  {ICLR} 2024, Vienna, Austria, May 7-11, 2024},
  publisher    = {OpenReview.net},
  year         = {2024},
  url          = {https://openreview.net/forum?id=oKn9c6ytLx},
  bibsource    = {dblp computer science bibliography, https://dblp.org}
}

@inproceedings{koh2024visualwebarenaEvaluatingMultimodal,
  author       = {Jing Yu Koh and
                  Robert Lo and
                  Lawrence Jang and
                  Vikram Duvvur and
                  Ming Chong Lim and
                  Po{-}Yu Huang and
                  Graham Neubig and
                  Shuyan Zhou and
                  Russ Salakhutdinov and
                  Daniel Fried},
  editor       = {Lun{-}Wei Ku and
                  Andre Martins and
                  Vivek Srikumar},
  title        = {VisualWebArena: Evaluating Multimodal Agents on Realistic Visual Web
                  Tasks},
  booktitle    = {Proceedings of the 62nd Annual Meeting of the Association for Computational
                  Linguistics (Volume 1: Long Papers), {ACL} 2024, Bangkok, Thailand,
                  August 11-16, 2024},
  pages        = {881--905},
  publisher    = {Association for Computational Linguistics},
  year         = {2024},
  url          = {https://doi.org/10.18653/v1/2024.acl-long.50},
  doi          = {10.18653/V1/2024.ACL-LONG.50},
  bibsource    = {dblp computer science bibliography, https://dblp.org}
}

@inproceedings{rawles2025androidworldDynamicBenchmarking,
  author       = {Christopher Rawles and
                  Sarah Clinckemaillie and
                  Yifan Chang and
                  Jonathan Waltz and
                  Gabrielle Lau and
                  Marybeth Fair and
                  Alice Li and
                  William E. Bishop and
                  Wei Li and
                  Folawiyo Campbell{-}Ajala and
                  Daniel Kenji Toyama and
                  Robert James Berry and
                  Divya Tyamagundlu and
                  Timothy P. Lillicrap and
                  Oriana Riva},
  title        = {AndroidWorld: {A} Dynamic Benchmarking Environment for Autonomous
                  Agents},
  booktitle    = {The Thirteenth International Conference on Learning Representations,
                  {ICLR} 2025, Singapore, April 24-28, 2025},
  publisher    = {OpenReview.net},
  year         = {2025},
  url          = {https://openreview.net/forum?id=il5yUQsrjC},
  bibsource    = {dblp computer science bibliography, https://dblp.org}
}

@inproceedings{xie2024osworldBenchmarkingMultimodal,
  author       = {Tianbao Xie and
                  Danyang Zhang and
                  Jixuan Chen and
                  Xiaochuan Li and
                  Siheng Zhao and
                  Ruisheng Cao and
                  Toh Jing Hua and
                  Zhoujun Cheng and
                  Dongchan Shin and
                  Fangyu Lei and
                  Yitao Liu and
                  Yiheng Xu and
                  Shuyan Zhou and
                  Silvio Savarese and
                  Caiming Xiong and
                  Victor Zhong and
                  Tao Yu},
  editor       = {Amir Globersons and
                  Lester Mackey and
                  Danielle Belgrave and
                  Angela Fan and
                  Ulrich Paquet and
                  Jakub M. Tomczak and
                  Cheng Zhang},
  title        = {OSWorld: Benchmarking Multimodal Agents for Open-Ended Tasks in Real
                  Computer Environments},
  booktitle    = {Advances in Neural Information Processing Systems 38: Annual Conference
                  on Neural Information Processing Systems 2024, NeurIPS 2024, Vancouver,
                  BC, Canada, December 10 - 15, 2024},
  year         = {2024},
  url          = {http://papers.nips.cc/paper\_files/paper/2024/hash/5d413e48f84dc61244b6be550f1cd8f5-Abstract-Datasets\_and\_Benchmarks\_Track.html},
  bibsource    = {dblp computer science bibliography, https://dblp.org}
}

@inproceedings{ouyang2025kernelbenchCanLlms,
  author       = {Anne Ouyang and
                  Simon Guo and
                  Simran Arora and
                  Alex L. Zhang and
                  William Hu and
                  Christopher R{\'{e}} and
                  Azalia Mirhoseini},
  editor       = {Aarti Singh and
                  Maryam Fazel and
                  Daniel Hsu and
                  Simon Lacoste{-}Julien and
                  Felix Berkenkamp and
                  Tegan Maharaj and
                  Kiri Wagstaff and
                  Jerry Zhu},
  title        = {KernelBench: Can LLMs Write Efficient {GPU} Kernels?},
  booktitle    = {Forty-second International Conference on Machine Learning, {ICML}
                  2025, Vancouver, BC, Canada, July 13-19, 2025},
  series       = {Proceedings of Machine Learning Research},
  publisher    = {{PMLR} / OpenReview.net},
  year         = {2025},
  url          = {https://proceedings.mlr.press/v267/ouyang25a.html},
  bibsource    = {dblp computer science bibliography, https://dblp.org}
}

@article{merrill2026terminal-benchBenchmarkingAgents,
  author       = {Mike A. Merrill and
                  Alexander Glenn Shaw and
                  Nicholas Carlini and
                  Boxuan Li and
                  Harsh Raj and
                  Ivan Bercovich and
                  Lin Shi and
                  Jeong Yeon Shin and
                  Thomas Walshe and
                  Estefany Kelly Buchanan and
                  Junhong Shen and
                  Guanghao Ye and
                  Haowei Lin and
                  Jason Poulos and
                  Maoyu Wang and
                  Marianna Nezhurina and
                  Jenia Jitsev and
                  Di Lu and
                  Orfeas Menis{-}Mastromichalakis and
                  Zhiwei Xu and
                  Zizhao Chen and
                  Yue Liu and
                  Robert Zhang and
                  Leon Liangyu Chen and
                  Anurag Kashyap and
                  Jan{-}Lucas Uslu and
                  Jeffrey Li and
                  Jianbo Wu and
                  Minghao Yan and
                  Song Bian and
                  Vedang Sharma and
                  Ke Sun and
                  Steven Dillmann and
                  Akshay Anand and
                  Andrew Lanpouthakoun and
                  Bardia Koopah and
                  Changran Hu and
                  Etash Kumar Guha and
                  Gabriel H. S. Dreiman and
                  Jiacheng Zhu and
                  Karl Krauth and
                  Li Zhong and
                  Niklas Muennighoff and
                  Robert Amanfu and
                  Shangyin Tan and
                  Shreyas Pimpalgaonkar and
                  Tushar Aggarwal and
                  Xiangning Lin and
                  Xin Lan and
                  Xuandong Zhao and
                  Yiqing Liang and
                  Yuanli Wang and
                  Zilong Wang and
                  Changzhi Zhou and
                  David Heineman and
                  Hange Liu and
                  Harsh Trivedi and
                  John Yang and
                  Junhong Lin and
                  Manish Shetty and
                  Michael Yang and
                  Nabil Omi and
                  Negin Raoof and
                  Shanda Li and
                  Terry Yue Zhuo and
                  Wuwei Lin and
                  Yiwei Dai and
                  Yuxin Wang and
                  Wenhao Chai and
                  Shang Zhou and
                  Dariush Wahdany and
                  Ziyu She and
                  Jiaming Hu and
                  Zhikang Dong and
                  Yuxuan Zhu and
                  Sasha Cui and
                  Ahson Saiyed and
                  Arinbj{\"{o}}rn Kolbeinsson and
                  Jesse Hu and
                  Christopher Michael Rytting and
                  Ryan Marten and
                  Yixin Wang and
                  Alex Dimakis and
                  Andy Konwinski and
                  Ludwig Schmidt},
  title        = {Terminal-Bench: Benchmarking Agents on Hard, Realistic Tasks in Command
                  Line Interfaces},
  journal      = {CoRR},
  volume       = {abs/2601.11868},
  year         = {2026},
  url          = {https://doi.org/10.48550/arXiv.2601.11868},
  doi          = {10.48550/ARXIV.2601.11868},
  eprinttype   = {arXiv},
  eprint       = {2601.11868},
  bibsource    = {dblp computer science bibliography, https://dblp.org}
}

@inproceedings{yang2025swe-benchMultimodalDo,
  author       = {John Yang and
                  Carlos E. Jimenez and
                  Alex L. Zhang and
                  Kilian Lieret and
                  Joyce Yang and
                  Xindi Wu and
                  Ori Press and
                  Niklas Muennighoff and
                  Gabriel Synnaeve and
                  Karthik R. Narasimhan and
                  Diyi Yang and
                  Sida Wang and
                  Ofir Press},
  title        = {SWE-bench Multimodal: Do {AI} Systems Generalize to Visual Software
                  Domains?},
  booktitle    = {The Thirteenth International Conference on Learning Representations,
                  {ICLR} 2025, Singapore, April 24-28, 2025},
  publisher    = {OpenReview.net},
  year         = {2025},
  url          = {https://openreview.net/forum?id=riTiq3i21b},
  bibsource    = {dblp computer science bibliography, https://dblp.org}
}

@article{wang2025ragenUnderstandingSelf-evolution,
  author       = {Zihan Wang and
                  Kangrui Wang and
                  Qineng Wang and
                  Pingyue Zhang and
                  Linjie Li and
                  Zhengyuan Yang and
                  Xing Jin and
                  Kefan Yu and
                  Minh Nhat Nguyen and
                  Licheng Liu and
                  Eli Gottlieb and
                  Yiping Lu and
                  Kyunghyun Cho and
                  Jiajun Wu and
                  Li Fei{-}Fei and
                  Lijuan Wang and
                  Yejin Choi and
                  Manling Li},
  title        = {{RAGEN:} Understanding Self-Evolution in {LLM} Agents via Multi-Turn
                  Reinforcement Learning},
  journal      = {CoRR},
  volume       = {abs/2504.20073},
  year         = {2025},
  url          = {https://doi.org/10.48550/arXiv.2504.20073},
  doi          = {10.48550/ARXIV.2504.20073},
  eprinttype   = {arXiv},
  eprint       = {2504.20073},
  bibsource    = {dblp computer science bibliography, https://dblp.org}
}

@article{feng2025group-in-groupPolicyOptimization,
  author       = {Lang Feng and
                  Zhenghai Xue and
                  Tingcong Liu and
                  Bo An},
  title        = {Group-in-Group Policy Optimization for {LLM} Agent Training},
  journal      = {CoRR},
  volume       = {abs/2505.10978},
  year         = {2025},
  url          = {https://doi.org/10.48550/arXiv.2505.10978},
  doi          = {10.48550/ARXIV.2505.10978},
  eprinttype   = {arXiv},
  eprint       = {2505.10978},
  bibsource    = {dblp computer science bibliography, https://dblp.org}
}

@article{jin2025search-r1TrainingLlms,
  author       = {Bowen Jin and
                  Hansi Zeng and
                  Zhenrui Yue and
                  Dong Wang and
                  Hamed Zamani and
                  Jiawei Han},
  title        = {Search-R1: Training LLMs to Reason and Leverage Search Engines with
                  Reinforcement Learning},
  journal      = {CoRR},
  volume       = {abs/2503.09516},
  year         = {2025},
  url          = {https://doi.org/10.48550/arXiv.2503.09516},
  doi          = {10.48550/ARXIV.2503.09516},
  eprinttype   = {arXiv},
  eprint       = {2503.09516},
  bibsource    = {dblp computer science bibliography, https://dblp.org}
}

@article{hu2026seeupo,
  author       = {Tianyi Hu and
                  Qingxu Fu and
                  Yanxi Chen and
                  Zhaoyang Liu and
                  Bolin Ding},
  title        = {SeeUPO: Sequence-Level Agentic-RL with Convergence Guarantees},
  journal      = {CoRR},
  volume       = {abs/2602.06554},
  year         = {2026},
  url          = {https://doi.org/10.48550/arXiv.2602.06554},
  doi          = {10.48550/ARXIV.2602.06554},
  eprinttype   = {arXiv},
  eprint       = {2602.06554},
  bibsource    = {dblp computer science bibliography, https://dblp.org}
}

@article{li2025websailor,
  author       = {Kuan Li and
                  Zhongwang Zhang and
                  Huifeng Yin and
                  Liwen Zhang and
                  Litu Ou and
                  Jialong Wu and
                  Wenbiao Yin and
                  Baixuan Li and
                  Zhengwei Tao and
                  Xinyu Wang and
                  Weizhou Shen and
                  Junkai Zhang and
                  Dingchu Zhang and
                  Xixi Wu and
                  Yong Jiang and
                  Ming Yan and
                  Pengjun Xie and
                  Fei Huang and
                  Jingren Zhou},
  title        = {WebSailor: Navigating Super-human Reasoning for Web Agent},
  journal      = {CoRR},
  volume       = {abs/2507.02592},
  year         = {2025},
  url          = {https://doi.org/10.48550/arXiv.2507.02592},
  doi          = {10.48550/ARXIV.2507.02592},
  eprinttype   = {arXiv},
  eprint       = {2507.02592},
  bibsource    = {dblp computer science bibliography, https://dblp.org}
}

@inproceedings{wei2025webagent-r1,
  author       = {Zhepei Wei and
                  Wenlin Yao and
                  Yao Liu and
                  Weizhi Zhang and
                  Qin Lu and
                  Liang Qiu and
                  Changlong Yu and
                  Puyang Xu and
                  Chao Zhang and
                  Bing Yin and
                  Hyokun Yun and
                  Lihong Li},
  editor       = {Christos Christodoulopoulos and
                  Tanmoy Chakraborty and
                  Carolyn Rose and
                  Violet Peng},
  title        = {WebAgent-R1: Training Web Agents via End-to-End Multi-Turn Reinforcement
                  Learning},
  booktitle    = {Proceedings of the 2025 Conference on Empirical Methods in Natural
                  Language Processing, {EMNLP} 2025, Suzhou, China, November 4-9, 2025},
  pages        = {7909--7928},
  publisher    = {Association for Computational Linguistics},
  year         = {2025},
  url          = {https://doi.org/10.18653/v1/2025.emnlp-main.401},
  doi          = {10.18653/V1/2025.EMNLP-MAIN.401},
  bibsource    = {dblp computer science bibliography, https://dblp.org}
}

@article{lu2025ui-s1,
  author       = {Zhengxi Lu and
                  Jiabo Ye and
                  Fei Tang and
                  Yongliang Shen and
                  Haiyang Xu and
                  Ziwei Zheng and
                  Weiming Lu and
                  Ming{-}Hsuan Yang and
                  Fei Huang and
                  Jun Xiao and
                  Yueting Zhuang},
  title        = {{UI-S1:} Advancing {GUI} Automation via Semi-online Reinforcement
                  Learning},
  journal      = {CoRR},
  volume       = {abs/2509.11543},
  year         = {2025},
  url          = {https://doi.org/10.48550/arXiv.2509.11543},
  doi          = {10.48550/ARXIV.2509.11543},
  eprinttype   = {arXiv},
  eprint       = {2509.11543},
  bibsource    = {dblp computer science bibliography, https://dblp.org}
}

@article{jin2025videomemEnhancingUltra-long,
  author       = {Hongbo Jin and
                  Qingyuan Wang and
                  Wenhao Zhang and
                  Yang Liu and
                  Sijie Cheng},
  title        = {VideoMem: Enhancing Ultra-Long Video Understanding via Adaptive Memory
                  Management},
  journal      = {CoRR},
  volume       = {abs/2512.04540},
  year         = {2025},
  url          = {https://doi.org/10.48550/arXiv.2512.04540},
  doi          = {10.48550/ARXIV.2512.04540},
  eprinttype   = {arXiv},
  eprint       = {2512.04540},
  bibsource    = {dblp computer science bibliography, https://dblp.org}
}

@article{liu2025gemGymAgentic,
  author       = {Zichen Liu and
                  Anya Sims and
                  Keyu Duan and
                  Changyu Chen and
                  Simon Yu and
                  Xiangxin Zhou and
                  Haotian Xu and
                  Shaopan Xiong and
                  Bo Liu and
                  Chenmien Tan and
                  Chuen Yang Beh and
                  Weixun Wang and
                  Hao Zhu and
                  Weiyan Shi and
                  Diyi Yang and
                  Michael Shieh and
                  Yee Whye Teh and
                  Wee Sun Lee and
                  Min Lin},
  title        = {{GEM:} {A} Gym for Agentic LLMs},
  journal      = {CoRR},
  volume       = {abs/2510.01051},
  year         = {2025},
  url          = {https://doi.org/10.48550/arXiv.2510.01051},
  doi          = {10.48550/ARXIV.2510.01051},
  eprinttype   = {arXiv},
  eprint       = {2510.01051},
  bibsource    = {dblp computer science bibliography, https://dblp.org}
}

@article{stojanovski2025reasoningGymReasoning,
  author       = {Zafir Stojanovski and
                  Oliver Stanley and
                  Joe Sharratt and
                  Richard Jones and
                  Abdulhakeem Adefioye and
                  Jean Kaddour and
                  Andreas K{\"{o}}pf},
  title        = {{REASONING} {GYM:} Reasoning Environments for Reinforcement Learning
                  with Verifiable Rewards},
  journal      = {CoRR},
  volume       = {abs/2505.24760},
  year         = {2025},
  url          = {https://doi.org/10.48550/arXiv.2505.24760},
  doi          = {10.48550/ARXIV.2505.24760},
  eprinttype   = {arXiv},
  eprint       = {2505.24760},
  bibsource    = {dblp computer science bibliography, https://dblp.org}
}

@misc{meng2026gymvunifiedvisionenvironment,
      title={Gym-V: A Unified Vision Environment System for Agentic Vision Research}, 
      author={Fanqing Meng and Lingxiao Du and Jiawei Gu and Jiaqi Liao and Linjie Li and Zijian Wu and Xiangyan Liu and Ziqi Zhao and Mengkang Hu and Zichen Liu and Jiaheng Zhang and Michael Qizhe Shieh},
      year={2026},
      eprint={2603.15432},
      archivePrefix={arXiv},
      primaryClass={cs.CV},
      url={https://arxiv.org/abs/2603.15432}, 
}

@misc{aggarwal2026gymanythingturnsoftwareagent,
      title={Gym-Anything: Turn any Software into an Agent Environment}, 
      author={Pranjal Aggarwal and Graham Neubig and Sean Welleck},
      year={2026},
      eprint={2604.06126},
      archivePrefix={arXiv},
      primaryClass={cs.LG},
      url={https://arxiv.org/abs/2604.06126}, 
}

@inproceedings{xu2024wizardlm,
  author       = {Can Xu and
                  Qingfeng Sun and
                  Kai Zheng and
                  Xiubo Geng and
                  Pu Zhao and
                  Jiazhan Feng and
                  Chongyang Tao and
                  Qingwei Lin and
                  Daxin Jiang},
  title        = {WizardLM: Empowering Large Pre-Trained Language Models to Follow Complex
                  Instructions},
  booktitle    = {The Twelfth International Conference on Learning Representations,
                  {ICLR} 2024, Vienna, Austria, May 7-11, 2024},
  publisher    = {OpenReview.net},
  year         = {2024},
  url          = {https://openreview.net/forum?id=CfXh93NDgH},
  bibsource    = {dblp computer science bibliography, https://dblp.org}
}

@inproceedings{wang2023self-instruct,
  author       = {Yizhong Wang and
                  Yeganeh Kordi and
                  Swaroop Mishra and
                  Alisa Liu and
                  Noah A. Smith and
                  Daniel Khashabi and
                  Hannaneh Hajishirzi},
  editor       = {Anna Rogers and
                  Jordan L. Boyd{-}Graber and
                  Naoaki Okazaki},
  title        = {Self-Instruct: Aligning Language Models with Self-Generated Instructions},
  booktitle    = {Proceedings of the 61st Annual Meeting of the Association for Computational
                  Linguistics (Volume 1: Long Papers), {ACL} 2023, Toronto, Canada,
                  July 9-14, 2023},
  pages        = {13484--13508},
  publisher    = {Association for Computational Linguistics},
  year         = {2023},
  url          = {https://doi.org/10.18653/v1/2023.acl-long.754},
  doi          = {10.18653/V1/2023.ACL-LONG.754},
  bibsource    = {dblp computer science bibliography, https://dblp.org}
}

@article{fang2025towardsGeneralAgentic,
  author       = {Runnan Fang and
                  Shihao Cai and
                  Baixuan Li and
                  Jialong Wu and
                  Guangyu Li and
                  Wenbiao Yin and
                  Xinyu Wang and
                  Xiaobin Wang and
                  Liangcai Su and
                  Zhen Zhang and
                  Shibin Wu and
                  Zhengwei Tao and
                  Yong Jiang and
                  Pengjun Xie and
                  Fei Huang and
                  Jingren Zhou},
  title        = {Towards General Agentic Intelligence via Environment Scaling},
  journal      = {CoRR},
  volume       = {abs/2509.13311},
  year         = {2025},
  url          = {https://doi.org/10.48550/arXiv.2509.13311},
  doi          = {10.48550/ARXIV.2509.13311},
  eprinttype   = {arXiv},
  eprint       = {2509.13311},
  bibsource    = {dblp computer science bibliography, https://dblp.org}
}

@article{song2026envscalerScalingTool-interactive,
  author       = {Xiaoshuai Song and
                  Haofei Chang and
                  Guanting Dong and
                  Yutao Zhu and
                  Zhicheng Dou and
                  Ji{-}Rong Wen},
  title        = {EnvScaler: Scaling Tool-Interactive Environments for {LLM} Agent via
                  Programmatic Synthesis},
  journal      = {CoRR},
  volume       = {abs/2601.05808},
  year         = {2026},
  url          = {https://doi.org/10.48550/arXiv.2601.05808},
  doi          = {10.48550/ARXIV.2601.05808},
  eprinttype   = {arXiv},
  eprint       = {2601.05808},
  bibsource    = {dblp computer science bibliography, https://dblp.org}
}

@article{tu2026scaleenvScalingEnvironment,
  author       = {Dunwei Tu and
                  Hongyan Hao and
                  Hansi Yang and
                  Yihao Chen and
                  Yi{-}Kai Zhang and
                  Zhikang Xia and
                  Yu Yang and
                  Yueqing Sun and
                  Xingchen Liu and
                  Furao Shen and
                  Qi Gu and
                  Hui Su and
                  Xunliang Cai},
  title        = {ScaleEnv: Scaling Environment Synthesis from Scratch for Generalist
                  Interactive Tool-Use Agent Training},
  journal      = {CoRR},
  volume       = {abs/2602.06820},
  year         = {2026},
  url          = {https://doi.org/10.48550/arXiv.2602.06820},
  doi          = {10.48550/ARXIV.2602.06820},
  eprinttype   = {arXiv},
  eprint       = {2602.06820},
  bibsource    = {dblp computer science bibliography, https://dblp.org}
}

@article{li2025questaExpandingReasoning,
  author       = {Jiazheng Li and
                  Hong Lu and
                  Kaiyue Wen and
                  Zaiwen Yang and
                  Jiaxuan Gao and
                  Hongzhou Lin and
                  Yi Wu and
                  Jingzhao Zhang},
  title        = {QuestA: Expanding Reasoning Capacity in LLMs via Question Augmentation},
  journal      = {CoRR},
  volume       = {abs/2507.13266},
  year         = {2025},
  url          = {https://doi.org/10.48550/arXiv.2507.13266},
  doi          = {10.48550/ARXIV.2507.13266},
  eprinttype   = {arXiv},
  eprint       = {2507.13266},
  bibsource    = {dblp computer science bibliography, https://dblp.org}
}

@article{zhang2025scaf-grpoScaffoldedGroup,
  author       = {Xichen Zhang and
                  Sitong Wu and
                  Yinghao Zhu and
                  Haoru Tan and
                  Shaozuo Yu and
                  Ziyi He and
                  Jiaya Jia},
  title        = {Scaf-GRPO: Scaffolded Group Relative Policy Optimization for Enhancing
                  {LLM} Reasoning},
  journal      = {CoRR},
  volume       = {abs/2510.19807},
  year         = {2025},
  url          = {https://doi.org/10.48550/arXiv.2510.19807},
  doi          = {10.48550/ARXIV.2510.19807},
  eprinttype   = {arXiv},
  eprint       = {2510.19807},
  bibsource    = {dblp computer science bibliography, https://dblp.org}
}

@article{zhang2025adhintAdaptiveHints,
  author       = {Feng Zhang and
                  Zezhong Tan and
                  Xinhong Ma and
                  Ziqiang Dong and
                  Xi Leng and
                  Jianfei Zhao and
                  Xin Sun and
                  Yang Yang},
  title        = {ADHint: Adaptive Hints with Difficulty Priors for Reinforcement Learning},
  journal      = {CoRR},
  volume       = {abs/2512.13095},
  year         = {2025},
  url          = {https://doi.org/10.48550/arXiv.2512.13095},
  doi          = {10.48550/ARXIV.2512.13095},
  eprinttype   = {arXiv},
  eprint       = {2512.13095},
  bibsource    = {dblp computer science bibliography, https://dblp.org}
}

@article{zhang2025stephintMulti-levelStepwise,
  author       = {Kaiyi Zhang and
                  Ang Lv and
                  Jinpeng Li and
                  Yongbo Wang and
                  Feng Wang and
                  Haoyuan Hu and
                  Rui Yan},
  title        = {StepHint: Multi-level Stepwise Hints Enhance Reinforcement Learning
                  to Reason},
  journal      = {CoRR},
  volume       = {abs/2507.02841},
  year         = {2025},
  url          = {https://doi.org/10.48550/arXiv.2507.02841},
  doi          = {10.48550/ARXIV.2507.02841},
  eprinttype   = {arXiv},
  eprint       = {2507.02841},
  bibsource    = {dblp computer science bibliography, https://dblp.org}
}

@article{huang2025boostingMllmReasoning,
  author       = {Qihan Huang and
                  Long Chan and
                  Jinlong Liu and
                  Wanggui He and
                  Hao Jiang and
                  Mingli Song and
                  Jingyuan Chen and
                  Chang Yao and
                  Jie Song},
  title        = {Boosting {MLLM} Reasoning with Text-Debiased Hint-GRPO},
  journal      = {CoRR},
  volume       = {abs/2503.23905},
  year         = {2025},
  url          = {https://doi.org/10.48550/arXiv.2503.23905},
  doi          = {10.48550/ARXIV.2503.23905},
  eprinttype   = {arXiv},
  eprint       = {2503.23905},
  bibsource    = {dblp computer science bibliography, https://dblp.org}
}

@article{kimi_k2,
  author       = {Kimi Team},
  title        = {Kimi {K2:} Open Agentic Intelligence},
  journal      = {CoRR},
  volume       = {abs/2507.20534},
  year         = {2025},
  url          = {https://doi.org/10.48550/arXiv.2507.20534},
  doi          = {10.48550/ARXIV.2507.20534},
  eprinttype   = {arXiv},
  eprint       = {2507.20534},
  bibsource    = {dblp computer science bibliography, https://dblp.org}
}

@article{environment_survey,
  author       = {Jiachun Li and
                  Zhuoran Jin and
                  Tianyi Men and
                  Yupu Hao and
                  Kejian Zhu and
                  Lingshuai Wang and
                  Dongqi Huang and
                  Longxiang Wang and
                  Shengjia Hua and
                  Lu Wang and
                  Jinshan Gao and
                  Hongbang Yuan and
                  Ruilin Xu and
                  Kang Liu and
                  Jun Zhao},
  title        = {Agentic Environment Engineering for Large Language Models: {A} Survey
                  of Environment Modeling, Synthesis, Evaluation, and Application},
  journal      = {CoRR},
  volume       = {abs/2606.12191},
  year         = {2026},
  url          = {https://doi.org/10.48550/arXiv.2606.12191},
  doi          = {10.48550/ARXIV.2606.12191},
  eprinttype   = {arXiv},
  eprint       = {2606.12191},
  bibsource    = {dblp computer science bibliography, https://dblp.org}
}

\clearpage
\appendix

\newpage

\section{RL Algorithms}
\label{appendix:rl_algorithms}

In this section, we elaborates on the specific formulations and optimization objectives of the RL algorithms.
We follow the notations in Section \ref{sec:preliminary}: for each goal $g$, the old policy $\pi_{\theta_{\mathrm{old}}}$ samples a group of $N$ trajectories $\{\tau_k\}_{k=1}^{N}$, and each trajectory receives a trajectory-level reward $r(\tau_k)$. Let $\mathbf{a}_k=(a_{k,1},\ldots,a_{k,L_k})$ denote the concatenation of all agent-generated tokens in $\tau_k$, where $L_k$ is the number of such tokens. The conditioning context for token $a_{k,t}$, including the goal, the observations, and the previous interaction history, is denoted by $c_{k,t}$. Environment-generated tokens are excluded from all policy-gradient sums below. Unless otherwise specified, the expectations are taken over goals and trajectory groups sampled from $\pi_{\theta_{\mathrm{old}}}$.

The group-normalized trajectory advantage is computed as
\begin{equation}
    A_k = A(\tau_k) = \frac{r(\tau_k)-\mathrm{mean}\left(\{r(\tau_j)\}_{j=1}^{N}\right)}{\mathrm{std}\left(\{r(\tau_j)\}_{j=1}^{N}\right)} ,
\end{equation}
and is assigned to every agent-generated token in the same trajectory, i.e., $A_{k,t}=A_k$.

\paragraph{GRPO.}
Group Relative Policy Optimization (GRPO) uses token-level importance ratios while estimating advantages from the relative rewards within the sampled group. Its clipped objective can be written as

{\small 
\begin{equation}
\mathcal{J}_{\mathrm{GRPO}}(\theta) = \mathbb{E} \Bigg[ \frac{1}{N}\sum_{k=1}^{N}\frac{1}{L_k}\sum_{t=1}^{L_k}
\min\left(
\rho_{k,t}(\theta) A_k,
\mathrm{clip}\big(\rho_{k,t}(\theta),1-\varepsilon,1+\varepsilon\big) A_k
\right) \Bigg],
\end{equation}
}

where
\begin{equation}
    \rho_{k,t}(\theta)=
    \frac{\pi_{\theta}(a_{k,t}\mid c_{k,t})}
    {\pi_{\theta_{\mathrm{old}}}(a_{k,t}\mid c_{k,t})} .
\end{equation}
Thus, GRPO performs clipping and policy-gradient weighting independently for each generated token, while the reward signal remains trajectory-level.

\paragraph{DAPO.}
Decoupled Clip and Dynamic Sampling Policy Optimization (DAPO) preserves the group-relative advantage estimation but modifies the GRPO reduction and clipping rule. In particular, it uses a token-level loss reduction over all agent-generated tokens in the group and decouples the lower and upper clipping ranges:

{\small
\begin{equation}
\label{eq:dapo_loss}
\mathcal{J}_{\mathrm{DAPO}}(\theta) = \mathbb{E} \Bigg[ \frac{1}{\sum_{k=1}^{N}L_k}\sum_{k=1}^{N}\sum_{t=1}^{L_k}
\min\left(
\rho_{k,t}(\theta) A_k,
\mathrm{clip}\big(\rho_{k,t}(\theta), 1-\varepsilon_{\mathrm{low}}, 1+\varepsilon_{\mathrm{high}}\big) A_k
\right) \Bigg],
\end{equation}
}
with the same token-level ratio $\rho_{k,t}(\theta)$ and group-normalized advantage $A_k$ as above. Dynamic sampling further keeps only non-degenerate trajectory groups. For binary success rewards, this condition is:
\begin{equation}
    0 < \left|\{\tau_k \mid r(\tau_k)=1\}\right| < N,
\end{equation}
or, more generally, groups whose rewards have non-zero variance. This filtering avoids batches in which all trajectories receive identical rewards and therefore produce zero normalized advantages.

\paragraph{GSPO.}
Group Sequence Policy Optimization (GSPO) instead aligns the optimization unit with the trajectory-level reward by defining the importance ratio at the sequence level. For each trajectory, the length-normalized sequence ratio over agent-generated tokens is defined as the following form:
\begin{equation}
s_k(\theta)
=\left(\prod_{t=1}^{L_k}\frac{\pi_{\theta}(a_{k,t}\mid c_{k,t})}{\pi_{\theta_{\mathrm{old}}}(a_{k,t}\mid c_{k,t})}\right)^{\frac{1}{L_k}}
=\exp\left(\frac{1}{L_k}\sum_{t=1}^{L_k}
\log\frac{\pi_{\theta}(a_{k,t}\mid c_{k,t})}
{\pi_{\theta_{\mathrm{old}}}(a_{k,t}\mid c_{k,t})}\right),
\end{equation}
GSPO optimizes the following objective:

{\small
\begin{equation}
\label{eq:gspo_loss}
\mathcal{J}_{\mathrm{GSPO}}(\theta) = \mathbb{E} \Bigg[ \frac{1}{N}\sum_{k=1}^{N}
\min\left(
s_k(\theta) A_k,
\mathrm{clip}\big(s_k(\theta),1-\varepsilon,1+\varepsilon\big) A_k
\right) \Bigg].
\end{equation}
}

The clipping decision is therefore made once per trajectory rather than once per token, while all agent-generated tokens in the trajectory share the same sequence-level weight and advantage.

\section{Enrichment Strategies}
\label{appendix:enrichment_strategies}

In this section, we detail the construction of enriched environments based on two standard environments, SciWorld and BFCL, following the given enrichment strategy. For each environment, we first introduce the basic rule-based implementation and then provide concrete reference examples.

\subsection{Enriched Environments for SciWorld}

For SciWorld, the implementation of action guidance leverages the ground-truth expert trajectories provided for each task. 
For observation enrichment we utilize the task progress tracking logic available in the SciWorld source code. SciWorld internally maintains rule-based checks for evaluating task completion conditions and intermediate progress. Based on these signals, we enrich the environment observations with supplementary state information reflecting the current task status, enabling the agent to better perceive hidden environment dynamics and long-horizon progress.  Representative examples of the enriched environments from SciWorld are shown in Table \ref{tab:Action_Guidance_in_SciWorld} and \ref{tab:Observation_Enrichment_in_SciWorld}. The implementation is based on RLVMR \footnote{https://github.com/Tencent/DigitalHuman/tree/main/RLVMR}.

\subsection{Enriched Environments for BFCL-V3}

For BFCL-V3, we construct enriched environments across four domains: \textit{Vehicle Control}, \textit{Trading Bots}, \textit{Travel Booking}, and \textit{Gorilla File System}. We manually designate Gorilla File System and Vehicle Control for training, while reserving Trading Bots and Travel Booking for held-out evaluation. The action guidance in BFCL is implemented by appending lightweight hints after each user query. For observation enrichment, we augment the outputs of selected tools with additional execution information. Concretely, rather than returning sparse or empty responses, enriched feedback may indicate whether a tool invocation has successfully completed, provide intermediate execution status, or expose supplementary contextual information relevant to the current interaction state. For example, in Gorilla File System, enriched feedback can include additional path-related information, while in Vehicle Control, tool outputs may be extended with status indicators or auxiliary environment details.
Representative examples of BFCL enrichments are shown in Table \ref{tab:Action_Guidance_in_BFCL_V3_1}, Table \ref{tab:Action_Guidance_in_BFCL_V3_2} and Table \ref{tab:observation_enrichment_in_bfcl_v3}. The implementation is based on verl-agent \footnote{https://github.com/langfengQ/verl-agent}.

\section{Internalization Experiments}
\label{appendix:internalization_exp}

In this section, we provide the probe questions constructed for the Gorilla File System operations used in Exp 3. The complete examples corresponding to \texttt{cd}, \texttt{mkdir}, \texttt{touch}, \texttt{rm}, \texttt{rmdir}, \texttt{mv}, \texttt{cp}, and \texttt{echo} are presented in Table~\ref{tab:probing_questions}. For each probe question, the model predicts between two candidate outputs: the original feedback option $P(A)$ and the enriched feedback option $P(B)$. A higher $P(A)$ indicates that the model remains biased toward standard tool outputs and shows limited memory of the enriched feedback, whereas a higher $P(B)$ suggests that the model recalls and internalizes the information introduced by the enriched feedback.

\begin{table*}[htbp]
    \footnotesize
    \centering
    \begin{tabular}{p{14cm}}
        \toprule
You are an expert agent operating in the ScienceWorld environment, which is a text-based virtual environment centered around accomplishing tasks from the elementary science curriculum.

\colorbox{blue!10}{Your current task is}: Your task is to determine whether round seed shape is a dominant or recessive trait in the unknown E plant. If the trait is dominant, focus on the green box. If the trait is recessive, focus on the blue box.

\colorbox{blue!10}{Your current observation is}: This room is called the greenhouse. In it, you see: 
	the agent
	a substance called air
	a bee hive. The bee hive door is closed. 
	a blue box (containing nothing)
	a flower pot 3 (containing soil)
	a flower pot 4 (containing soil)
	a flower pot 5 (containing soil)
	a flower pot 6 (containing soil)
	a flower pot 8 (containing soil)
	a flower pot 9 (containing soil)
	a green box (containing nothing)
	a jug (containing nothing)
	a seed jar (containing a square blue unknown E seed, a round blue unknown E seed)
	a sink, which is turned off. In the sink is: nothing.
You also see:
	A door to the hallway (that is open)
	A door to the outside (that is open)
Here are the actions you may take:

[

{"action": "open OBJ", "description": "open a container"},

{"action": "close OBJ", "description": "close a container"},

{"action": "activate OBJ", "description": "activate a device"},

{"action": "deactivate OBJ", "description": "deactivate a device"},

{"action": "connect OBJ to OBJ", "description": "connect electrical components"},

{"action": "disconnect OBJ", "description": "disconnect electrical components"},

{"action": "use OBJ [on OBJ]", "description": "use a device/item"},

{"action": "look around", "description": "describe the current room"},

{"action": "look at OBJ", "description": "describe an object in detail"},

{"action": "look in OBJ", "description": "describe a container's contents"},

{"action": "read OBJ", "description": "read a note or book"},

{"action": "move OBJ to OBJ", "description": "move an object to a container"},

{"action": "pick up OBJ", "description": "move an object to the inventory"},

{"action": "put down OBJ", "description": "drop an inventory item"},

{"action": "pour OBJ into OBJ", "description": "pour a liquid into a container"},

{"action": "dunk OBJ into OBJ", "description": "dunk a container into a liquid"},

{"action": "mix OBJ", "description": "chemically mix a container"},

{"action": "go to LOC", "description": "move to a new location"},

{"action": "eat OBJ", "description": "eat a food"},

{"action": "flush OBJ", "description": "flush a toilet"},

{"action": "focus on OBJ", "description": "signal intent on a task object"},

{"action": "wait", "description": "take no action for 10 iterations"},

{"action": "wait1", "description": "take no action for 1 iteration"},

{"action": "task", "description": "describe current task"},

{"action": "inventory", "description": "list your inventory"}

]

\colorbox{blue!10}{Current available actions}:

Valid\_actions: ['activate OBJ', 'close OBJ', 'deactivate OBJ', 'dunk OBJ in OBJ', 'eat OBJ', 'flush OBJ', 'focus on OBJ', 'go OBJ', 'inventory', 'look around', 'look at OBJ', 'look in OBJ', 'mix OBJ', 'move OBJ to OBJ', 'open OBJ', 'pick up OBJ', 'pour OBJ in OBJ', 'put down OBJ', 'read OBJ', 'reset task', 'task', 'teleport OBJ', 'use OBJ on OBJ', 'wait', 'wait1'], OBJ needs to be replaced with one of the following objects: ['agent', 'air', 'bee hive', 'blue box', 'ceramic cup', 'door to hallway', 'door to outside', 'flower pot 3', 'flower pot 4', 'flower pot 5', 'flower pot 6', 'flower pot 8', 'flower pot 9', 'green box', 'greenhouse', 'hallway', 'jug', 'outside', 'round blue unknown e seed', 'seed square blue unknown e seed', 'sink', 'soil in flower pot 3', 'soil in flower pot 4', 'soil in flower pot 5', 'soil in flower pot 6', 'soil in flower pot 8', 'soil in flower pot 9']

example: <action>focus on door</action>

\colorbox{blue!10}{Now it's your turn to take an action.}

You should first reason step-by-step about the current situation. This reasoning process MUST be enclosed within <think> </think> tags. 

Once you've finished your reasoning, you should choose an appropriate action for the current step and present it within <action> </action> tags.

{ \color{blue}

[Hint] Start with the following action to start your exploration!

['look around', 'pick up seed jar', 'look around']

}

\\
    \bottomrule
    \end{tabular}
    \caption{Action Guidance in SciWorld}
    \label{tab:Action_Guidance_in_SciWorld}
\end{table*}

\begin{table*}[htbp]
    \footnotesize
    \centering
    \begin{tabular}{p{14cm}}
        \toprule

You are an expert agent operating in the ScienceWorld environment, which is a text-based virtual environment centered around accomplishing tasks from the elementary science curriculum.

\colorbox{blue!10}{Your current task is}: Your task is to find a(n) non-living thing. First, focus on the thing. Then, move it to the yellow box in the living room.

Prior to this step, you have already taken 6 step(s). 

\colorbox{blue!10}{Below are the most recent 2 observations and the corresponding actions you took:}

[Step 1, Action 1: 'go to living room']

[Step 2, Action 2: 'focus on yellow box']

[Step 3, Action 3: 'pick up non-living thing']

[Step 4, Action 4: 'pick up object']

[Step 5, Observation 5: 'You move the pillow to the inventory.', Action 5: 'move pillow to yellow box']

[Step 6, Observation 6: 'You move the pillow to the yellow box.', Action 6: 'reset task']

\colorbox{blue!10}{You are now at step 7 and your current observation is:}

You reset the goal progress and focus.

Here are the actions you may take:

[

{"action": "open OBJ", "description": "open a container"},

{"action": "close OBJ", "description": "close a container"},

{"action": "activate OBJ", "description": "activate a device"},

{"action": "deactivate OBJ", "description": "deactivate a device"},

{"action": "connect OBJ to OBJ", "description": "connect electrical components"},

{"action": "disconnect OBJ", "description": "disconnect electrical components"},

{"action": "use OBJ [on OBJ]", "description": "use a device/item"},

{"action": "look around", "description": "describe the current room"},

{"action": "look at OBJ", "description": "describe an object in detail"},

{"action": "look in OBJ", "description": "describe a container's contents"},

{"action": "read OBJ", "description": "read a note or book"},

{"action": "move OBJ to OBJ", "description": "move an object to a container"},

{"action": "pick up OBJ", "description": "move an object to the inventory"},

{"action": "put down OBJ", "description": "drop an inventory item"},

{"action": "pour OBJ into OBJ", "description": "pour a liquid into a container"},

{"action": "dunk OBJ into OBJ", "description": "dunk a container into a liquid"},

{"action": "mix OBJ", "description": "chemically mix a container"},

{"action": "go to LOC", "description": "move to a new location"},

{"action": "eat OBJ", "description": "eat a food"},

{"action": "flush OBJ", "description": "flush a toilet"},

{"action": "focus on OBJ", "description": "signal intent on a task object"},

{"action": "wait", "description": "take no action for 10 iterations"},

{"action": "wait1", "description": "take no action for 1 iteration"},

{"action": "task", "description": "describe current task"},

{"action": "inventory", "description": "list your inventory"}

]

\colorbox{blue!10}{Current available actions:}

Valid\_actions: ['activate OBJ', 'close OBJ', 'deactivate OBJ', 'dunk OBJ in OBJ', 'eat OBJ', 'flush OBJ', 'focus on OBJ', 'go OBJ', 'inventory', 'look around', 'look at OBJ', 'look in OBJ', 'mix OBJ', 'move OBJ to OBJ', 'open OBJ', 'pick up OBJ', 'pour OBJ in OBJ', 'put down OBJ', 'read OBJ', 'reset task', 'task', 'teleport OBJ', 'use OBJ on OBJ', 'wait', 'wait1'], OBJ needs to be replaced with one of the following objects: ['agent', 'air', 'book shelf', 'chair', 'cloth sittable', 'door', 'hallway', 'living room', 'object', 'painting', 'steel table', 'yellow box']
 example: <action>focus on door</action>

\colorbox{blue!10}{Now it's your turn to take an action.}

You should first reason step-by-step about the current situation. This reasoning process MUST be enclosed within <think> </think> tags.

Once you've finished your reasoning, you should choose an appropriate action for the current step and present it within <action> </action> tags.

{ \color{blue}
[Hint] Considering The current goal progress is Completed keys: 
\begin{verbatim}
Sequential Subgoals:
0	false	               GoalFocusOnNonlivingThing	focus on a non-living thing
1	false	                   GoalObjectInContainer	move living thing to answer box
Unordered and Optional Subgoals:
0	false	                   GoalObjectInContainer	Pick up object
1	true	                  GoalInRoomWithOpenDoor	Be in a room with an open door
2	true	                   GoalMoveToNewLocation	Move to a new location
3	true	                      GoalMoveToLocation	Move to the location of the answer box
\end{verbatim}
}

\\
    \bottomrule
    \end{tabular}
    \caption{Observation Enrichment in SciWorld}
    \label{tab:Observation_Enrichment_in_SciWorld}

\end{table*}

\begin{table*}[htbp]
    \footnotesize
    \centering
    \begin{tabular}{p{14cm}}
        \toprule

\textcolor{system}{\textbf{[System]}} You are an expert in composing functions.You are given a question and a set of possible functions. Based on the question, you will need to make one or more function/tool calls to achieve the purpose. If none of the functions can be used, point it out. If the given question lacks the parameters required by the function, also point it out.

You should only return the function calls in your response.

If you decide to invoke any of the function(s), you MUST put it in the format of [ func\_name1(params\_name1=params\_value1, params\_name2=params\_value2...), func\_name2(params)].  You SHOULD NOT include any other text in the response.

At each turn, you should try your best to complete the tasks requested by the user within the current turn. Continue to output functions to call until you have fulfilled the user's request to the best of your ability. Once you have no more functions to call, the system will consider the current turn complete and proceed to the next turn or task.

Here is a list of functions in json format that you can invoke.

......

\vspace{0.8em}

\noindent
\textcolor{user}{\textbf{[User]}} Find a file named 'config.py' somewhere deep in the file system and once you have located it, display the last line of the first occurring file.

{\color{blue}
Hint: considering start with the following operation: cd(folder='projects')
} 

\vspace{0.8em}

\noindent
\textcolor{assistant}{\textbf{[Assistant]}} [cd(folder='projects'), find(path='.', name='config.py'), tail(file\_name='config.py', lines=1)]

\vspace{0.8em}

\noindent
\textcolor{tool}{\textbf{[Tool]}} \{"current\_working\_directory": "projects"\}

\vspace{0.8em}

\noindent
\textcolor{tool}{\textbf{[Tool]}} \{"matches": ["./deep\_folder/config.py", "./deep\_folder/real\_config.py"]\}

\vspace{0.8em}

\noindent
\textcolor{tool}{\textbf{[Tool]}} \{"error": "tail: config.py: No such file or directory"\}

\vspace{0.8em}

\\
    \bottomrule
    \end{tabular}
        \caption{Action Guidance in BFCL-V3}
    \label{tab:Action_Guidance_in_BFCL_V3_1}

\end{table*}

\begin{table*}[htbp]
    \footnotesize
    \centering
    \begin{tabular}{p{14cm}}
        \toprule

\textcolor{system}{\textbf{[System]}} You are an expert in composing functions.You are given a question and a set of possible functions. Based on the question, you will need to make one or more function/tool calls to achieve the purpose. If none of the functions can be used, point it out. If the given question lacks the parameters required by the function, also point it out.

You should only return the function calls in your response.

If you decide to invoke any of the function(s), you MUST put it in the format of [ func\_name1(params\_name1=params\_value1, params\_name2=params\_value2...), func\_name2(params)].  You SHOULD NOT include any other text in the response.

At each turn, you should try your best to complete the tasks requested by the user within the current turn. Continue to output functions to call until you have fulfilled the user's request to the best of your ability. Once you have no more functions to call, the system will consider the current turn complete and proceed to the next turn or task.

Here is a list of functions in json format that you can invoke.

......

\vspace{0.8em}

\noindent
\textcolor{user}{\textbf{[User]}}I need you to set up a fresh folder wherever you're currently working.

{\color{blue}
 Hint: Essential information is missing from the user's request. Do not make any assumptions; explicitly ask the user to clarify the missing parameters.
} 

\vspace{0.8em}

\noindent
\textcolor{assistant}{\textbf{[Assistant]}} The user's request is missing the name of the folder to create. Please provide the name of the folder.

\vspace{0.8em}
\textcolor{user}{\textbf{[User]}} The fresh folder should name 'WebDevProjects'. 

\vspace{0.8em}
\textcolor{assistant}{\textbf{[Assistant]}}  [mkdir(dir\_name='WebDevProjects')]

\vspace{0.8em}
\noindent
\textcolor{tool}{\textbf{[Tool]}} None

\vspace{0.8em}
\textcolor{assistant}{\textbf{[Assistant]}}  The directory 'WebDevProjects' has been successfully created. Is there anything else you need assistance with?

\\
    \bottomrule
    \end{tabular}
        \caption{Action Guidance in BFCL-V3}
    \label{tab:Action_Guidance_in_BFCL_V3_2}

\end{table*}

\begin{table*}[htbp]
    \footnotesize
    \centering
    \begin{tabular}{p{14cm}}
        \toprule

\textcolor{user}{\textbf{[User]}} I'm cruising down a country road in my brand-new convertible, admiring the scenic lavender fields when I notice the fuel gauge creeping towards empty. I'd appreciate it if you could refill with 10 liters of gasoline to keep the adventure alive. Use 2 decimal digit of the gallon amount

\vspace{0.8em}
\textcolor{assistant}{\textbf{[Assistant]}}  [gallon\_to\_liter(liter=10.00)]

\vspace{0.8em}
\noindent
\textcolor{tool}{\textbf{[Tool]}} Error: VehicleControlAPI.gallon\_to\_liter() got an unexpected keyword argument 'liter'

\vspace{0.8em}
\textcolor{assistant}{\textbf{[Assistant]}}  [liter\_to\_gallon(liter=10.00)]

\vspace{0.8em}
\noindent
\textcolor{tool}{\textbf{[Tool]}} {"gallon": 2.6417200000000003}

\vspace{0.8em}
\textcolor{assistant}{\textbf{[Assistant]}}  [fillFuelTank(fuelAmount=2.64)]

\vspace{0.8em}
\noindent
\textcolor{tool}{\textbf{[Tool]}} {"fuelLevel": 4.640000000000001}

\vspace{0.8em}
\textcolor{assistant}{\textbf{[Assistant]}}  []

\vspace{0.8em}
\textcolor{user}{\textbf{[User]}} Now that the tank is replenished, let's fire up the engine with a swift ignition and take a peek at the dashboard stats, particularly the battery voltage and the AC temperature before we continue our picturesque journey.

\vspace{0.8em}
\textcolor{assistant}{\textbf{[Assistant]}}  [startEngine(ignitionMode="START")]

\vspace{0.8em}
\noindent
\textcolor{tool}{\textbf{[Tool]}} {"error": "All doors must be locked before starting the engine. Here are the unlocked doors: driver, passenger, rear\_left, rear\_right", "\_ignition\_diagnostic": {"all\_doors\_locked": false, "unlocked\_list": ["driver", "passenger", "rear\_left", "rear\_right"], "brake\_pedal\_position": 0.0, "fuel\_level\_gallons": 4.640000000000001, "battery\_voltage": 12.6}, "\_hint": "Ensure doors are locked, brake is pressed to 1.0, and fuel is > 0."}

\vspace{0.8em}
\textcolor{assistant}{\textbf{[Assistant]}}  [lockDoors(unlock=False, door=["driver", "passenger", "rear\_left", "rear\_right"])]

\vspace{0.8em}
\noindent
\textcolor{tool}{\textbf{[Tool]}} {"lockStatus": "locked", "remainingUnlockedDoors": 0, "\_current\_door\_grid": {"driver": "locked", "passenger": "locked", "rear\_left": "locked", "rear\_right": "locked"}, "\_summary": "0 doors currently unlocked."}

\vspace{0.8em}
\textcolor{assistant}{\textbf{[Assistant]}}  [startEngine(ignitionMode="START")]

\vspace{0.8em}
\noindent
\textcolor{tool}{\textbf{[Tool]}} {"error": "Brake pedal needs to be pressed when starting the engine.", "\_ignition\_diagnostic": {"all\_doors\_locked": true, "unlocked\_list": [], "brake\_pedal\_position": 0.0, "fuel\_level\_gallons": 4.640000000000001, "battery\_voltage": 12.6}, "\_hint": "Ensure doors are locked, brake is pressed to 1.0, and fuel is > 0."}

\vspace{0.8em}
\textcolor{assistant}{\textbf{[Assistant]}}  [pressBrakePedal(pedalPosition=1)]

\vspace{0.8em}
\noindent
\textcolor{tool}{\textbf{[Tool]}} {"brakePedalStatus": "pressed", "brakePedalForce": 1000.0}

\vspace{0.8em}
\textcolor{assistant}{\textbf{[Assistant]}}  [startEngine(ignitionMode="START")]

\vspace{0.8em}
\noindent
\textcolor{tool}{\textbf{[Tool]}} {"engineState": "running", "fuelLevel": 4.640000000000001, "batteryVoltage": 12.6}

\vspace{0.8em}
\textcolor{assistant}{\textbf{[Assistant]}}  [displayCarStatus(option="climate")]

\vspace{0.8em}
\noindent
\textcolor{tool}{\textbf{[Tool]}} {"currentACTemperature": 25.0, "fanSpeed": 50, "climateMode": "auto", "humidityLevel": 50.0}

\vspace{0.8em}
\textcolor{assistant}{\textbf{[Assistant]}}  []

\vspace{0.8em}
\textcolor{user}{\textbf{[User]}} Could you check the current status of a specific part of my vehicle and let me know?

\vspace{0.8em}
\textcolor{assistant}{\textbf{[Assistant]}}  [displayCarStatus(option="brakePadle")]

\vspace{0.8em}
\noindent
\textcolor{tool}{\textbf{[Tool]}} {"error": "Invalid option"}

\vspace{0.8em}
\textcolor{assistant}{\textbf{[Assistant]}}  [displayCarStatus(option="brakePedal")]

\vspace{0.8em}
\noindent
\textcolor{tool}{\textbf{[Tool]}} {"brakePedalStatus": "pressed", "brakePedalForce": 1000}

\vspace{0.8em}
\textcolor{assistant}{\textbf{[Assistant]}}  []

\vspace{0.8em}
\textcolor{user}{\textbf{[User]}} I'm keen on ensuring the tires' health is up to standard before we venture further into the rolling hills. Can you show me the tire pressure for each one?

\vspace{0.8em}
\textcolor{assistant}{\textbf{[Assistant]}}  [check\_tire\_pressure()]

\vspace{0.8em}
\noindent
\textcolor{tool}{\textbf{[Tool]}} {"frontLeftTirePressure": 32.0, "frontRightTirePressure": 32.0, "rearLeftTirePressure": 30.0, "rearRightTirePressure": 30.0, "healthy\_tire\_pressure": true, "car\_info": \{\}, \color{blue}{"\_decision\_safety\_hint": "NOTICE: 'healthy\_tire\_pressure' is based on the factory standard (30-35 psi). "}}

\vspace{0.8em}
......
\\
    \bottomrule
    \end{tabular}
        \caption{Observation Enrichement in BFCL-V3}
    \label{tab:observation_enrichment_in_bfcl_v3}

\end{table*}

\begin{table*}[htbp]
    \footnotesize
    \centering
    \begin{tabular}{p{14cm}}
        \toprule

{\tiny
\begin{verbatim}
PROBE_QUESTIONS = [
    {
        "name": "Q1_cd_bfcl_history",
        "question": "You have just been trained in the BFCL environment. Based on your recent training history, 
        when you successfully execute the tool `cd(folder='../src')`, 
        does the environment's return value include the `_new_directory_contents` information?",
        "options": {
            "A": "No, it does not include `_new_directory_contents`.",
            "B": "Yes, it automatically includes `_new_directory_contents`."
        }
    },
    {
        "name": "Q2_mkdir_bfcl_history",
        "question": "You have just been trained in the BFCL environment. Based on your recent training history, 
        when you successfully execute the tool `mkdir(dir_name='test_dir')`,
        does the environment's return value include the `_new_directory_contents` information?",
        "options": {
            "A": "No, it does not include `_new_directory_contents`.",
            "B": "Yes, it automatically includes `_new_directory_contents`."
        }
    },
    {
        "name": "Q3_touch_bfcl_history",
        "question": "You have just been trained in the BFCL environment. 
        Based on your recent training history, 
        when you successfully execute the tool `touch(file_name='app.log')`, 
        does the environment's return value include the `_new_directory_contents` information?",
        "options": {
            "A": "No, it does not include `_new_directory_contents`.",
            "B": "Yes, it automatically includes `_new_directory_contents`."
        }
    },
    {
        "name": "Q4_rm_bfcl_history",
        "question": "You have just been trained in the BFCL environment. 
        Based on your recent training history, 
        when you successfully execute the tool `rm(file_name='obsolete.txt')`, 
        does the environment's return value include the `_new_directory_contents` information?",
        "options": {
            "A": "No, it does not include `_new_directory_contents`.",
            "B": "Yes, it automatically includes `_new_directory_contents`."
        }
    },
    {
        "name": "Q5_rmdir_bfcl_history",
        "question": "You have just been trained in the BFCL environment. 
        Based on your recent training history, 
        when you successfully execute the tool `rmdir(dir_name='old_folder')`, 
        does the environment's return value include the `_new_directory_contents` information?",
        "options": {
            "A": "No, it does not include `_new_directory_contents`.",
            "B": "Yes, it automatically includes `_new_directory_contents`."
        }
    },
    {
        "name": "Q6_mv_bfcl_history",
        "question": "You have just been trained in the BFCL environment. 
        Based on your recent training history, 
        when you successfully execute the tool `mv(source='v1.py', destination='v2.py')`, 
        does the environment's return value include the `_new_directory_contents` information?",
        "options": {
            "A": "No, it does not include `_new_directory_contents`.",
            "B": "Yes, it automatically includes `_new_directory_contents`."
        }
    },
    {
        "name": "Q7_cp_bfcl_history",
        "question": "You have just been trained in the BFCL environment. 
        Based on your recent training history, 
        when you successfully execute the tool `cp(source='data.txt', destination='backup.txt')`,
        does the environment's return value include the `_new_directory_contents` information?",
        "options": {
            "A": "No, it does not include `_new_directory_contents`.",
            "B": "Yes, it automatically includes `_new_directory_contents`."
        }
    },
    {
        "name": "Q8_echo_bfcl_history",
        "question": "You have just been trained in the BFCL environment. 
        Based on your recent training history, 
        when you successfully execute the tool `echo(content='data', file_name='data.csv')` to write a file,
        does the environment's return value include the `_new_directory_contents` information?",
        "options": {
            "A": "No, it does not include `_new_directory_contents`.",
            "B": "Yes, it automatically includes `_new_directory_contents`."
        }
    }
]

\end{verbatim}
}
\\
    \bottomrule
    \end{tabular}

    \caption{Probing Questions in Research Question 3}
        \label{tab:probing_questions}
\end{table*}

\end{document}